\pdfoutput=1
\documentclass[lettersize,journal]{IEEEtran}
\usepackage{amsmath,amsfonts}
\usepackage{algorithmic}
\usepackage{algorithm}
\usepackage{array}
\usepackage[caption=false,font=normalsize,labelfont=sf,textfont=sf]{subfig}
\usepackage{textcomp}
\usepackage{stfloats}
\usepackage{url}
\usepackage{verbatim}
\usepackage{graphicx}
\usepackage{cite}

\usepackage[utf8]{inputenc} 
\usepackage[T1]{fontenc}    
\usepackage{hyperref}       
\usepackage{url}            
\usepackage{booktabs}       
\usepackage{amsfonts}       
\usepackage{nicefrac}       
\usepackage{microtype}      
\usepackage{xcolor}         
\usepackage{amsmath,amssymb,amsthm}
\usepackage{algorithm,algorithmic}
\usepackage{mathtools}
\usepackage[capitalize,noabbrev]{cleveref}
\usepackage{bbm}
\usepackage{bm}
\usepackage{array}
\usepackage{graphicx}
\usepackage{wrapfig} 
\usepackage{subfig}
\usepackage{multirow}
\usepackage{stfloats}
\usepackage{enumitem}

\theoremstyle{plain}
\newtheorem{theorem}{Theorem}[section]

\theoremstyle{definition}

\theoremstyle{remark}

\begin{document}

\title{Angel or Devil: Discriminating \\ Hard Samples and Anomaly Contaminations for \\ Unsupervised Time Series Anomaly Detection}

\author{Ruyi Zhang, Hongzuo Xu, Songlei Jian, Yusong Tan, Haifang Zhou, Rulin Xu
\thanks{Ruyi Zhang, Songlei Jian, Yusong Tan, Haifang Zhou, Rulin Xu are with the College of Computer, National University of Defense Technology, Changsha 410073, PR China. 
E-mail: \{zhangruyi, jiansonglei, ystan, haifang\_zhou, xurulin11\}@ nudt.edu.cn}
\thanks{Hongzuo Xu is with the 
Intelligent Game and Decision Lab, Beijing 100091, PR China.
E-mail: leogarcia@126.com}
\thanks{Hongzuo Xu and Songlei Jian are corresponding authors.}
}

\markboth{Preprint for IEEE Transactions on Knowledge and Data Engineering}%
{Shell \MakeLowercase{\textit{et al.}}: A Sample Article Using IEEEtran.cls for IEEE Journals}


\maketitle

\begin{abstract}
Training in unsupervised time series anomaly detection is constantly plagued by the discrimination between harmful \textit{anomaly contaminations} and beneficial \textit{hard normal samples}. These two samples exhibit analogous loss behavior that conventional loss-based methodologies struggle to differentiate. To tackle this problem, we propose a novel approach that supplements traditional loss behavior with \textit{parameter behavior}, enabling a more granular characterization of anomalous patterns. Parameter behavior is formalized by measuring the parametric response to minute perturbations in input samples. Leveraging the complementary nature of parameter and loss behaviors, we further propose a dual Parameter-Loss Data Augmentation method (termed PLDA), implemented within the reinforcement learning paradigm. During the training phase of anomaly detection, PLDA dynamically augments the training data through an iterative process that simultaneously mitigates anomaly contaminations while amplifying informative hard normal samples. PLDA demonstrates remarkable versatility, which can serve as an additional component that seamlessly integrated with existing anomaly detectors to enhance their detection performance. Extensive experiments on ten datasets show that PLDA significantly improves the performance of four distinct detectors by up to 8\%, outperforming three state-of-the-art data augmentation methods.
 
\end{abstract}


\section{Introduction}

Time series anomaly detection (TSAD) aims to detect data that deviate from expected normal patterns in continuous systems, with wide-ranging applications in fields like healthcare, finance, and transportation~\cite{lai2021revisiting,chalapathy2019deep}.
Due to the challenging and costly data labeling process, unsupervised TSAD is the mainstream of current research~\cite{braei2020anomaly,pang2021deep,darban2022deep}.
The key concept of unsupervised TSAD lies in accurately discerning normal patterns in the training set to detect anomalies in the testing set that differ from these patterns~\cite{chandola2009anomaly}. 
But this relies on the assumption of a pristine training set devoid of anomalies, which is often violated~\cite{han2020survey,kieu2022robust,du2021gan,li2022robust}.
Real-world training sets commonly contain unknown anomalies, termed anomaly contaminations (AC)~\cite{li2022learning,xu2023rosas,pang2023deep,niu2023graph,xu2024calibrated}.
Unthinkingly utilizing contaminated training sets can distort latent representations, leading to anomaly overfitting~\cite{kieu2022robust}. The overfitting can complicate anomaly detection by potentially assigning low scores to certain anomalies in the testing set.

\begin{figure}[!t]
    \centering
    \includegraphics[width=0.48\textwidth]{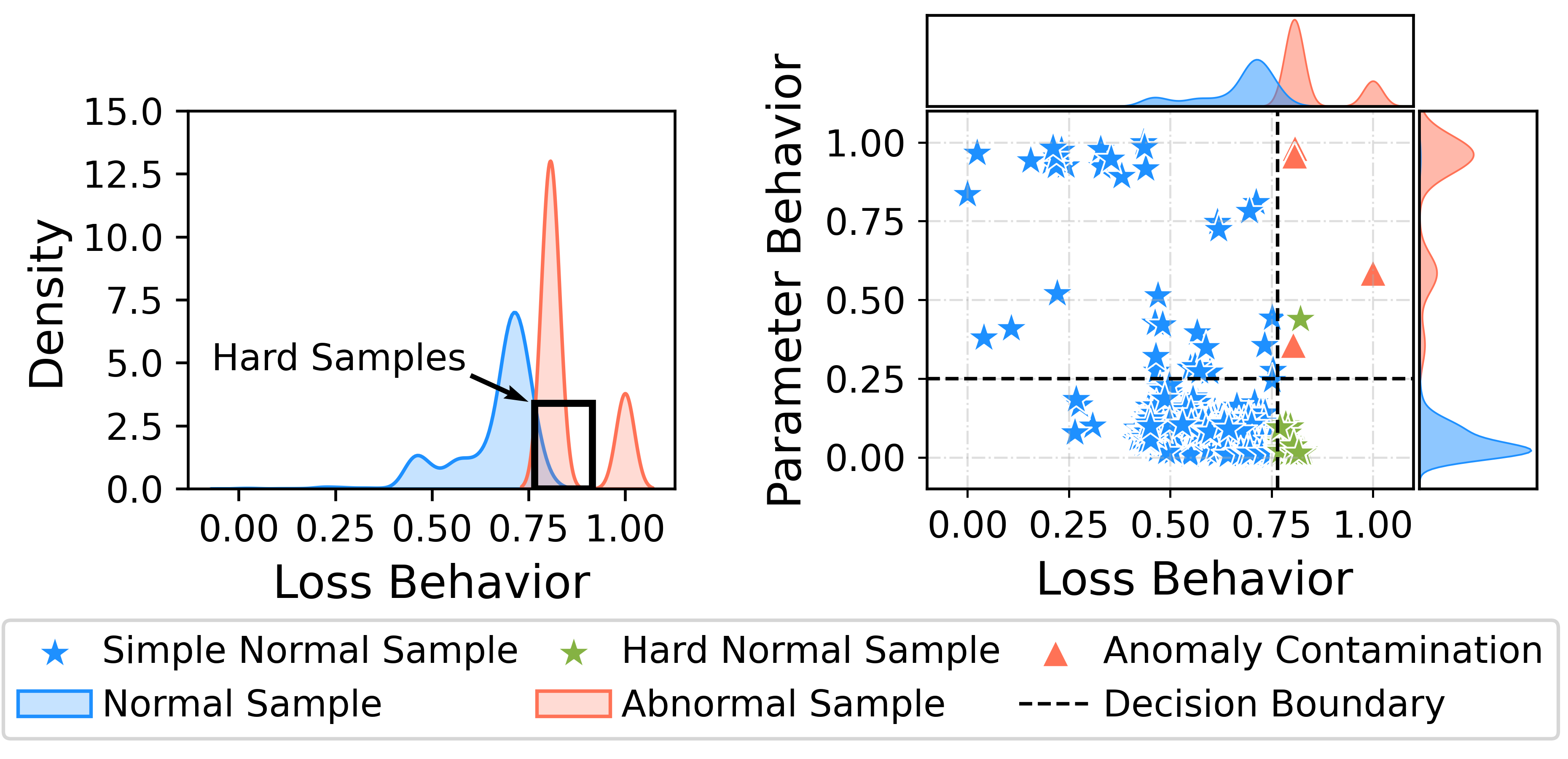} 
    \caption{Hard normal sample analysis.
        Loss value alone fails to differentiate the hard normal samples, but adding a new dimension, i.e., parameter behavior, allows for clear discrimination of sample types.}
    \label{fig:loss}
\end{figure}

Handling unsupervised TSAD in a contaminated training set meets a primary gap: \textit{The discrimination of anomaly contaminations and hard samples.} 
Hard-to-learn normal samples, also known as hard samples (HS), are normal samples located close to the decision boundary, which act as angels in clarifying the normal patterns~\cite{arpit2017closer,wu2018blockdrop,kishida2019empirical}. On the contrary, AC act as devils, which can severely damage the learned normal patterns. It is thus vital to keep more HS while reducing AC. Unfortunately, empirical observations show that AC and HS exhibit similarities in certain characteristics like large loss~\cite{mahsa2023differences,chang2017active}. 
Current studies widely utilize the small-loss trick to discern AC~(large loss) from normal samples~(small loss)~\cite{li2022robust,song2022learning,arazo2019unsupervised,li2019gradient}. 
As depicted in \cref{fig:loss}~(a), the single loss value often mistakenly discriminates between AC and HS.
Therefore, we intend to introduce a new dimension to express abnormal behavior, thus further discriminating between AC and HS. 

In TSAD training, loss $l=L(\mathbf{s}, \theta)$
is the mapping result of the data samples $\mathbf{s}$ under a set of parameters $\theta$, which leads to a coarse-grained reflection on the data. Instead, the fluctuations in parameters $\theta$ can reflect the subtle changes in the data at a finer granularity. Therefore, we introduce the parameter behavior alongside the loss behavior to describe the anomaly behavior.
To formalize the parameter behavior for a specific sample, it is natural to compare the parameters between the presence and absence of this sample. However, retraining the model for every data sample to get the results is prohibitively costly. 
Inspired by the works~\cite{hampel1974influence,koh2017understanding}, we evaluate how the TSAD model's parameters respond when a sample is slightly disturbed, i.e., the parameter sensitivity. 
As shown in \cref{fig:loss}~(b), 
we use the parameter sensitivity to quantify the parameter behavior and construct a dual-dimensional metric with loss behavior to 
help differentiate AC and HS.

Since AC ubiquitously exist in various time series, creating a new TSAD model may lack generality. We hope to develop a data augmentation plugin method that integrates with different TSAD models for broader applicability.
To comprehensively utilize both parameter and loss behavior, we propose a dual Parameter-Loss Data Augmentation method (PLDA for short) for unsupervised TSAD in contamination.
PLDA can iteratively identify and address AC and HS during the TSAD model's training phase.
Specifically, it employs a double DQN-based reinforcement learning paradigm for data augmentation. The augmentation is achieved by guiding an agent to select actions that maximize expected total future rewards. 
The dual-dimensional reward function considers parameter and loss behaviors to identify AC and HS effectively.
Additionally, we leverage the continuous nature of time series data and design an adaptive sliding window module as the action space, targeting AC reduction and HS enrichment.
PLDA can be employed as an additional step in the anomaly detector training process, providing the detector with a cleaner training set that helps it learn normal patterns better.

The main contributions are highlighted as follows:
\begin{itemize}
    \item We build the parameter behavior based on the parameter sensitivity and propose a dual parameter-loss data augmentation method, i.e., PLDA, which can iteratively identify and address AC and HS in the training set.
    \item We propose a reinforcement learning paradigm to implement PLDA, which leverages the parameter and loss behavior as a dual-dimensional reward. PLDA is model-independent and can be seamlessly incorporated into deep TSAD backbones.
    \item We formalize the impact of a data sample on the TSAD model as the parameter behavior and provide a theoretical analysis of its efficacy in identifying sample types.
\end{itemize}


Extensive experiments on ten datasets show that: (1) PLDA achieves a significant improvement in F1-scores compared to three state-of-the-art data augmentation methods; (2) PLDA enhances TSAD models, ensuring robust performance across different contamination rates. Additionally, we empirically validate the effectiveness of the proposed parameter behavior, dual-dimensional reward, and adaptive sliding window module in identifying and addressing AC and HS.



\section{Related Work}\label{sec:RelatedWork}
\textbf{Unsupervised Time Series Anomaly Detection.}~~
Due to the rarity and concealed nature of anomalies within a large volume of normal data points, accurately labeling data is both challenging and costly~\cite{xu2021anomaly}. As a result, the primary focus of recent research has shifted towards unsupervised TSAD~\cite{braei2020anomaly,pang2021deep}. Methods based on one-class classification~\cite{liu2008isolation,ruff2018deep} involve modeling the distribution of normal data, with any observations deviating from this distribution being identified as anomalies. Clustering-based methods~\cite{ghezelbash2020optimization,li2021clustering} typically consider observations that are significantly distant from any clustering centers as anomalies. Reconstruction-based methods~\cite{zhang2019deep,campos2021unsupervised} aim to reconstruct data from a low-dimensional space to capture essential normal patterns, operating under the assumption that anomalies cannot be accurately reconstructed. Overall, the fundamental concept of unsupervised TSAD is to learn the normal pattern or data distribution, which requires a clean training set devoid of any anomalies.



\textbf{Solutions for Deep TSAD in Contamination.}~~Recently, TSAD in contamination prompts numerous explorations.
Kieu~et~al. employ robust PCA for noise-signal data decomposition~\cite{kieu2022robust}. 
Some VAEs-based methods apply techniques that deal better with anomalies, such as heavy-tailed distributions~\cite{li2022learning} or $\alpha$, $\beta$, $\gamma$-divergences~\cite{kieu2022anomaly}. Some work adjusts the loss function based on the clustering result between the data~\cite{castellani2021estimating}. It relies on surface feature disparities, overlooking the underlying insights of their differences during the anomaly detection process.
Other solutions focus on filtering anomalies from the training set. Du~et~al. create pseudo-labels using the reconstruction error during training~\cite{du2021gan}, potentially deepening the model's "bias" against AC. Leveraging the memorization effect, other researchers employ the small-loss trick to filter anomalies~\cite{song2022learning}.
Li~et~al.~\cite{li2022robust} propose discarding samples with either high loss values or rapid changes in the loss. This method solely focuses on loss variations but overlooks the broader parameter space.

\textbf{Solutions for Other Fields in Contamination.}~~
Training set contamination, or noise labeling, also gains notable attention in other fields. 
Liu~et~al. compute the likelihood of data being clean using intra-class and inter-class similarity~\cite{liu2021noise}. This class feature-based method may not suit the imbalanced binary classification problem i.e., anomaly detection. 
Some studies also apply the small-loss trick to refine their loss functions. Arozo~et~al. employ a two-component beta mixture model on training loss to identify clean and noisy data~\cite{arazo2019unsupervised}, while Li~et~al. use a loss density metric to describe differing data loss distributions~\cite{li2019gradient}. These approaches separate noisy data by using a loss-related metric, yet they struggle to differentiate HS. 
Xia~et~al. leverage the parameter gradients and restrict the model to update only the key parameters with significant gradients each epoch~\cite{xia2020robust}. This work provids a valuable lesson in utilizing gradients to mitigate AC but potentially diminishing model sensitivity to HS. 
Some studies claim to retain HS.
Bai and Liu utilize the small-loss trick to find the confident normal data, and then they alternately update the confident samples and refine the classifier~\cite{bai2021me}. This method overemphasizes simple normal data and fails to reuse HS.
Zhang et al. calculate the prediction history over multiple training and set a threshold for HS ~\cite{zhang2022combating}. This off-line method cannot be iterated according to the training process of the model.

Current research neglects the parameters, focusing solely on loss values, whose granularity is too coarse to identify HS. Moreover, these studies typically rely on a singular metric to pinpoint AC, missing out on the benefits of integrating multiple metrics that could help differentiate HS through their complementary relationships.


\section{Parameter Behavior Modeling}\label{sec:ParamBehav}



\subsection{The Parameter Behavior Function}\label{sec:derivation}
In this section, we formalize the impact of a sample on the model as the parameter behavior by evaluating how the TSAD model's parameters respond when a sample is slightly disturbed.

Given a TSAD model with a parameter space $\Theta$, let $L(\mathbf{s}, \theta)$ be the loss function of a sample $\mathbf{s}$ under parameters $\theta\in\Theta$. Set $\hat{\theta}$ as the optimized parameters:
\begin{equation}\label{eq:hattheta}
    \begin{aligned}
    \hat{\theta} &= \mathop{\arg\min}\limits_{\theta}\frac{1}{n} \sum\limits_{i=1}^{n}L(\mathbf{s}_i, \theta).
    \end{aligned}
\end{equation}

Disturbing a sample $\mathbf{s}$ with a small weight $\epsilon$, we get the optimized parameters $\hat{\theta}_{\epsilon,\mathbf{s}}$:
\begin{equation}
\label{eq:theta}
\hat{\theta}_{\epsilon,\mathbf{s}} = \mathop{\arg\min}\limits_{\theta}\left[\frac{1}{n} \sum\limits_{i=1}^{n}L(\mathbf{s}_i, \theta)+\epsilon L(\mathbf{s},\theta)\right].
\end{equation}
We present the parameter sensitivity in \cref{the:dtheta} (proof in \cref{app:theA}).

\begin{theorem}\label{the:dtheta}
When a training sample $\mathbf{s}$ is disturbed with a small weight $\epsilon$, the gradient of the optimal parameter $\hat{\theta}_{\epsilon,\mathbf{s}}$ with respect to $\epsilon$ (i.e., parameter sensitivity) is:
    \begin{align}
     \left.\frac{\mathrm{d}\hat{\theta}_{\epsilon,\mathbf{s}}}{\mathrm{d}\epsilon}\right|_{\epsilon=0}=&-H^{-1}_{\hat{\theta}}\nabla_\theta L(\mathbf{s}, \hat{\theta}),
    \end{align} 
in which $H_{\hat{\theta}}=\frac{1}{n} \sum\limits_{i=1}^{n}\nabla_{\theta}^{2}L(\mathbf{s}_i, \theta)$ is the Hessian matrix.
\end{theorem}

As the distribution could be positive or negative, we define our parameter behavior function as:
\begin{equation}
    P(\mathbf{s}, \theta)=\left|H^{-1}_{\theta}\nabla_\theta L(\mathbf{s}, \theta)\right|.
\end{equation}

In practical applications, we utilize only the key parameters to compute the behavior value, thereby reducing the consumption of computational resources. These key parameters are defined as the $top$-$k$ parameters, which exhibit the most significant average behavior values during the first epoch.


\subsection{Effectiveness Analysis}\label{sec:effectInAC}
In this section, we employ Fourier transform~\cite{bochner1949fourier} to prove our parameter behavior's effectiveness in discriminating AC.
Fourier transform decomposes signals into periodic functions, allowing us to view a sample's parameter behavior as the sum of values across different frequency components:
\begin{equation}
    P(\mathbf{s}, \theta) = \sum\limits_{f} P(\mathbf{s}(f), \theta).
\end{equation}
Then we only need to focus on the behavior of parameter $\theta_j$ on each frequency $f$ component of the data $\mathbf{s}(f)$. We give the  \cref{the:lt} (proof in \cref{app:theB}).

\begin{theorem}\label{the:lt}
The absolute contribution from frequency $f$ to this total amount at parameter $\theta_j$ is:
\begin{equation}
    P(\mathbf{s}(f), \theta_j)=\left|\frac{\partial L(\mathbf{s}(f),\theta)}{\partial\theta_j}\right|\approx A(f)\mathrm{e}^{-\left|\frac{\pi f}{2\omega_j}\right|},
\end{equation}
in which $A(f)$ indicates the amplitude. 
\end{theorem}

\begin{figure}
\centering
\includegraphics[width=0.45\textwidth]{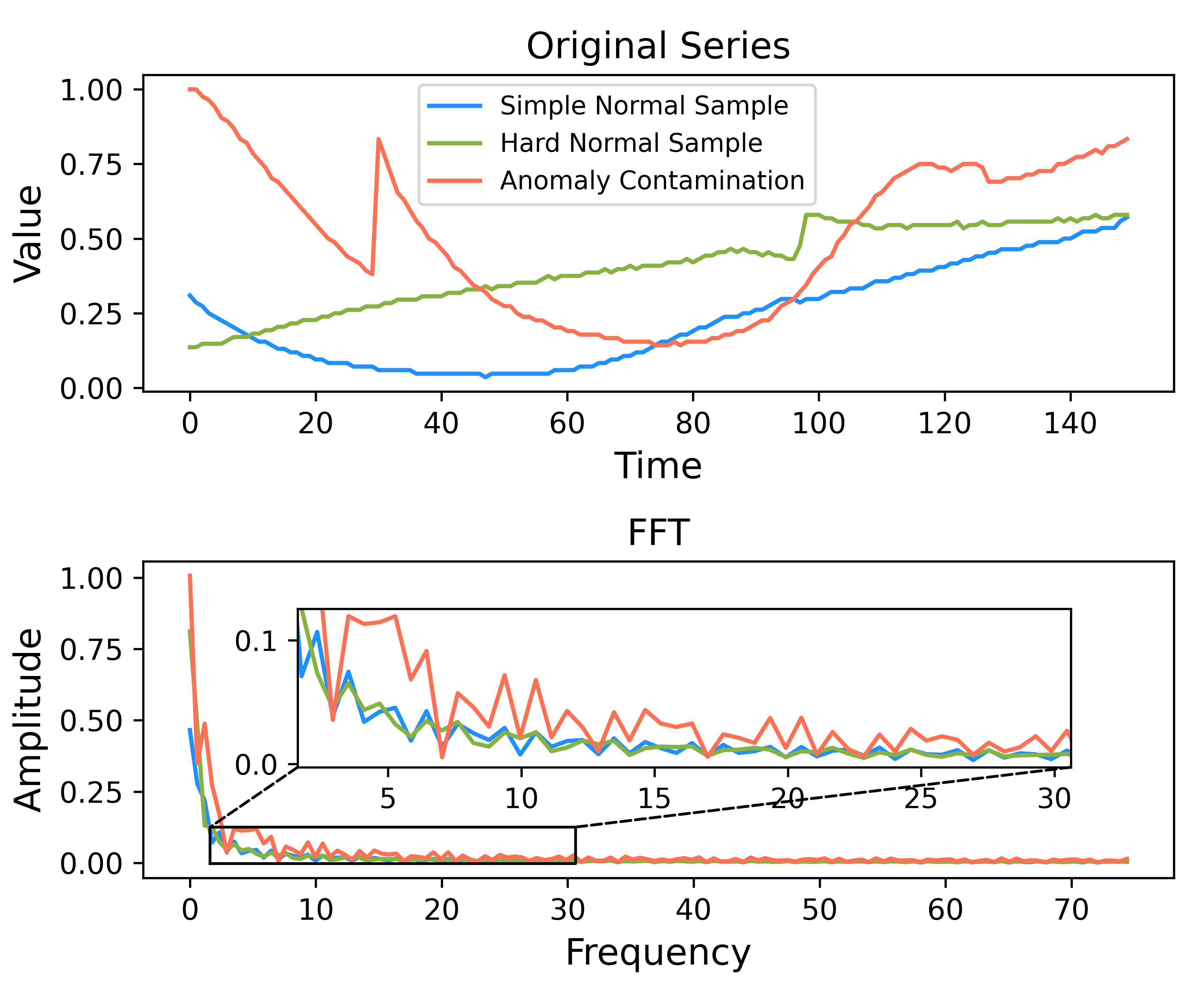}
\caption{The spectrograms of simple normal sample, hard normal sample, and anomaly contamination. Anomaly contamination contains more high-frequency components due to additional noise or abrupt changes. Hard normal sample also has some high-frequency components, but they are less pronounced than those in anomaly contamination.} \label{fig:fft}
\end{figure}

Based on \cref{the:lt}, neural networks tend to learn the lower frequency components of data preferentially. In other words, there is an inverse relationship between parameter behavior and data frequency. As shown in~\cref{fig:fft}, normal data and AC have different spectral characteristics, with AC showing more high-frequency components due to noise or abrupt changes~\cite{nam2024breaking,wang2024revisiting}. HS also contain high-frequency components, but these components are less pronounced than in AC. 
The prevalence of high-frequency components in AC results in distinct parameter behavior patterns compared to normal data. Our method exploits this distinction to effectively differentiate data categories.

\section{Method}\label{sec:Method}

%

In this section, we detail the implementation of PLDA's key modules and demonstrate how PLDA interacts with the TSAD model to augment the training set during the training phase iteratively. In the real-world implementation, the calculations are performed in batches to improve computation speed. To simplify the expression, we take a single sample as a state in this paper, for example.

\subsection{Problem Formulation}
Let $\mathcal{X}=\langle\mathbf{x}^0, \mathbf{x}^1, \ldots, \mathbf{x}^N\rangle$ be a time series with $N$ observations $\mathbf{x}\in\mathbbm{R}^D$, where $D$ denotes the feature dimension. 
The classic sliding window approach usually transforms $\mathcal{X}$ into a series of samples $\mathcal{S}=\{\mathbf{s}^0, \mathbf{s}^h, \mathbf{s}^{2h}, \ldots\}$, where each $\mathbf{s}^i=\langle \mathbf{x}^i, \mathbf{x}^{i+1}, \ldots, \mathbf{x}^{i+w-1}\rangle$ forms a time series sample starting at $\mathbf{x}^i$ with length $w$, using a window size $w$ and stride $h$.
The goal of TSAD is to train a model $M_\theta$, with $\theta$ representing the parameters, to capture normal patterns in the training set and detect anomalies in the testing set. Meanwhile, the goal of PLDA is to reduce AC and enrich HS in the training set, enhancing $M_\theta$'s ability to learn the normal patterns. \cref{tab:notation} summarises main notations.

\begin{table}[tb]
	\renewcommand{\arraystretch}{1.3}
	\caption{The main notations used in the paper}\label{tab:notation}
	\centering
	\scalebox{1}{
		\begin{tabular}{p{2.5cm}p{5.2cm}}
			\toprule[1.5pt]
			\textbf{Format}        & \textbf{Notations \mbox{-} Descriptions} \\
			\hline
			Calligraphic fonts     &  $\mathcal{X}$ \mbox{-} dataset, $\mathcal{S}$ \mbox{-} sub-sequence dataset                                \\
			Bold lowercase letters & $\mathbf{x}$ \mbox{-} data point, $\mathbf{s}$ \mbox{-} sample, 
			\newline$\mathbf{p}$ \mbox{-} parameter behavior value                             \\
			Lowercase letters & $w$ \mbox{-} window size, $h$ \mbox{-} stride, 
			\newline$r$ \mbox{-} reward value, $a$ \mbox{-} action, 
			\newline$e$ \mbox{-} number of data augmentation epoch,
			\newline$k$ \mbox{-} number of key parameters,
			\newline$f$ \mbox{-} data frequency
			\\
			Uppercase letters & $M$ \mbox{-} anomaly detection model,  
			\newline$L$ \mbox{-} loss function,      
			\newline$P$ \mbox{-} parameter behavior function,     
			\newline$Q$ \mbox{-} action-value function, 
			\newline$G$ \mbox{-} state transition function,
			\newline$R$ \mbox{-} reward function,
			\newline$H$ \mbox{-} Hessian matrix, $A$ \mbox{-} amplitude      
			\\
			Greek letters & $\theta$ \mbox{-} parameters of anomaly detection model, 
			\newline$\varphi$ \mbox{-} parameters of action-value function,
			\newline$\gamma$ \mbox{-} discount factor, $\alpha$ \mbox{-} balance factor,
			\newline$\phi$ \mbox{-} phase  \\
			
			\bottomrule[1.5pt]
	\end{tabular}}
\end{table}

\begin{figure*}[!t]
\captionsetup[subfloat]{captionskip=5pt, font=scriptsize }
\centering
\subfloat[PLDA]{\includegraphics[height=6.5cm]{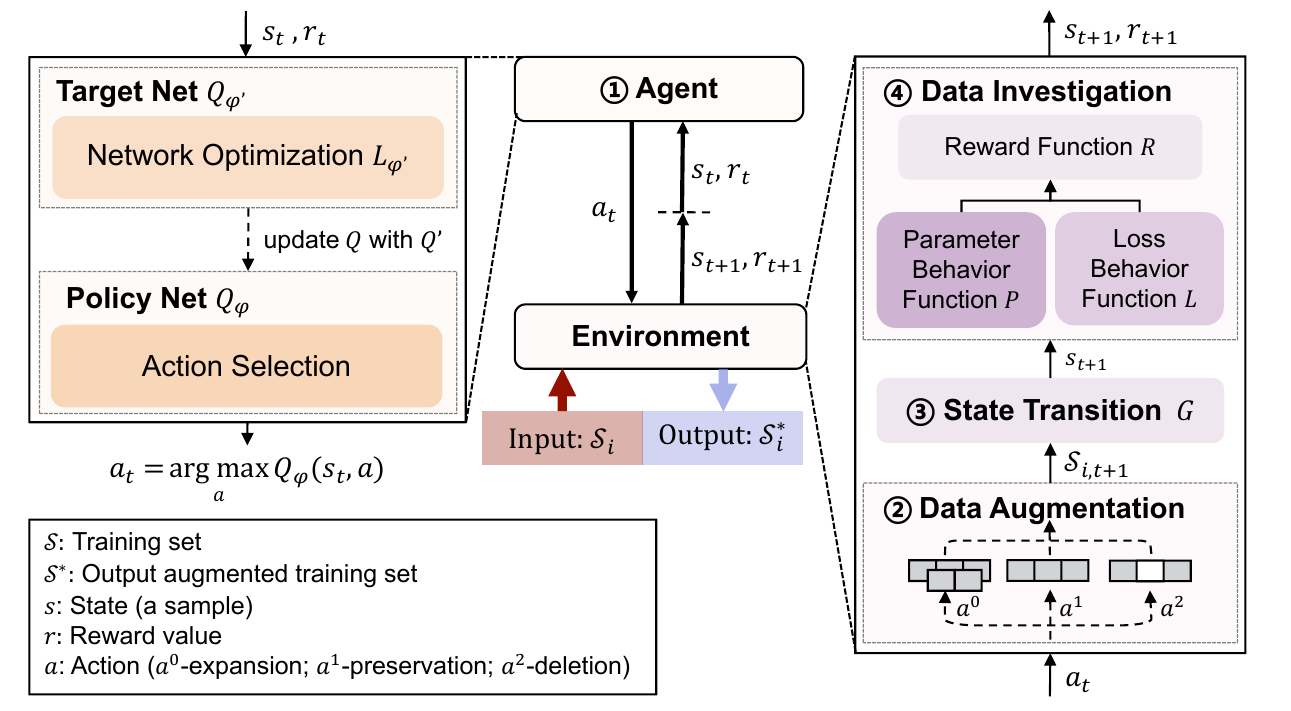}\label{fig:frameworka}}
\subfloat[Workflow]{\includegraphics[height=6.5cm]{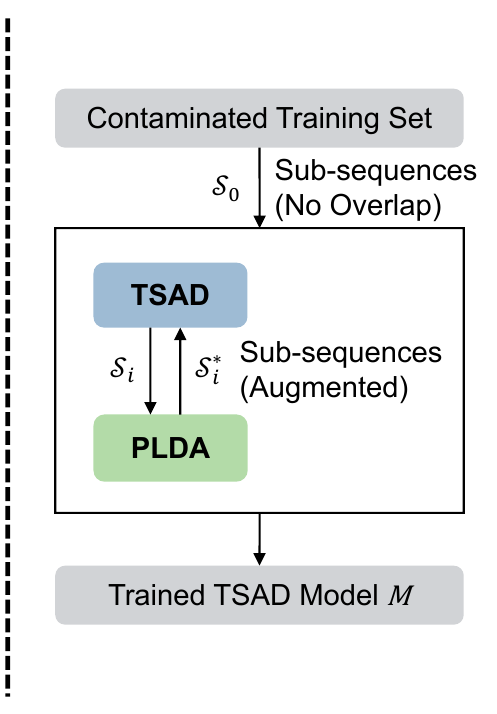}\label{fig:frameworkb}}
\caption{
Overview of PLDA. \cref{fig:frameworkb} illustrates the training workflow of TSAD, where PLDA acts as an additional component that iteratively augments the contaminated training set $\mathcal{S}_i$.
\cref{fig:frameworka} shows the augmentation details. Specifically, in $t$-th iteration, PLDA augments the state $\mathbf{s}_t$ sampled from $\mathcal{S}_i$ as follows: (1)~Agent selects optimal action $a_t$ that maximum expected total future reward. (2)~Data augmentation applies the chosen action, resulting in the augmented training set $\mathcal{S}_{i,t+1}$.
(3)~State transition function $G$ generates next state $\mathbf{s}_{t+1}\in\mathcal{S}_{i,t+1}$ for analysis. (4)~Data investigation computes the dual-dimensional reward $r_{t+1}$ of $\mathbf{s}_{t+1}$ based on parameter and loss behavior.
} 
\label{fig:framework}
\end{figure*}

\subsection{Overview}
The PLDA framework is illustrated in \cref{fig:framework}. PLDA works as an additional step in TSAD's training to enhance detection performance. In the $i$-th epoch of TSAD training, PLDA inputs the current training set $\mathcal{S}_{i}$ and outputs the augmented training set $\mathcal{S}_{i}^*$. Specifically, PLDA encompasses two core components: the agent and the environment. 
In each iteration, the agent selects the optimal action for the current state (i.e., a sample) according to the expected total future reward. The environment augments the training set with the optimal action, transforms it to the next state, and evaluates the dual-dimensional reward.
This sequence repeats, yielding the augmented $\mathcal{S}_i^*$.


We restructure the data augmentation into a deep reinforcement learning problem under the following four key modules: 
(1)~\textbf{Agent.} The agent is designed for data augmentation to determine an optimal action $a^*$ for the given state (i.e., a sample) $s_t$. The action space is represented by the set $\{a^0, a^1, a^2\}$, which represents expansion, preservation, and deletion operations, respectively. 
(2)~\textbf{Data augmentation.} The data augmentation adjusts the sampling rate of different data types by controlling the sliding window stride according to $a^*$.
In this way, the training set $\mathcal{S}_{i,t}$ is enhanced to $\mathcal{S}_{i,t+1}$.
(3)~\textbf{State transition.}
The state transition selects the next state $\mathbf{s}_{t+1}$ to be analyzed according to either randomness or predefined rules. The predefined rules ensure that critical data, such as AC and HS, are more likely to be chosen by imposing appropriate constraints.
(4)~\textbf{Data investigation.} The data investigation computes the dual-dimensional reward $r_{t+1}$ for the state $s_{t+1}$ based on the parameter behavior and loss behavior. The reward provides varying feedback depending on the type of action performed and the data type involved.

\subsection{Agent}\label{sec:agent}

Our agent aims to learn an action-value function, defined as the expected total future rewards that take action $a$ in the current state $\mathbf{s}$. The action value function can be approximated as follows:
\begin{equation}
    Q(\mathbf{s},a) = \mathbb{E}(r_t + \gamma r_{t+1} + \gamma^2 r_{t+2} + \cdots | \mathbf{s}_t=\mathbf{s}, a_t=a),
\end{equation}
in which $\gamma\in[0,1]$ is a discount factor to balance present and future rewards.  
To avoid overestimating the action value, we use a well-known double DQN~\cite{van2016deep} to learn $Q(\mathbf{s},a)$. 
This involves two networks: a policy net $Q_\varphi(\mathbf{s},a)$ for optimal action finding and a target net $Q_{\varphi'}(\mathbf{s},a)$ to assess current state value. 
These two networks share the same MLP architecture but different parameters~(i.e., $\varphi$ and $\varphi'$). $\varphi'$ is updated each iteration to minimize the following loss:
\begin{equation}\label{eq:varphi}
    L_{\varphi'} = \Vert r_t + \gamma\mathop{\max}\limits_{a_{t+1}}Q_{\varphi'}(\mathbf{s}_{t+1},a_{t+1})-Q_\varphi(\mathbf{s}_t,a_t) \Vert.
\end{equation}
In PLDA, $\varphi'$ is optimized using mini-batches sampled from memory $\{(s_i,a_i,r_i,s_{i+1})\}_{i=0}^{m}$, which is initially filled via warm start and subsequently refreshed each iteration. $\varphi$ is updated by $\varphi'$ every $q$ iterations. The optimal action is the one that maximizes the action value:
\begin{equation}
    a^* = \mathop{\arg\max}\limits_{a}Q_\varphi(\mathbf{s},a).
\end{equation}
\subsection{Environment}\label{sec:environment}

\subsubsection{Data Augmentation}
\begin{figure*}[!t]
\captionsetup[subfloat]{captionskip=5pt, font=scriptsize } 
\centering
\subfloat[Classic Sliding Window]{\includegraphics[height=4.5cm]{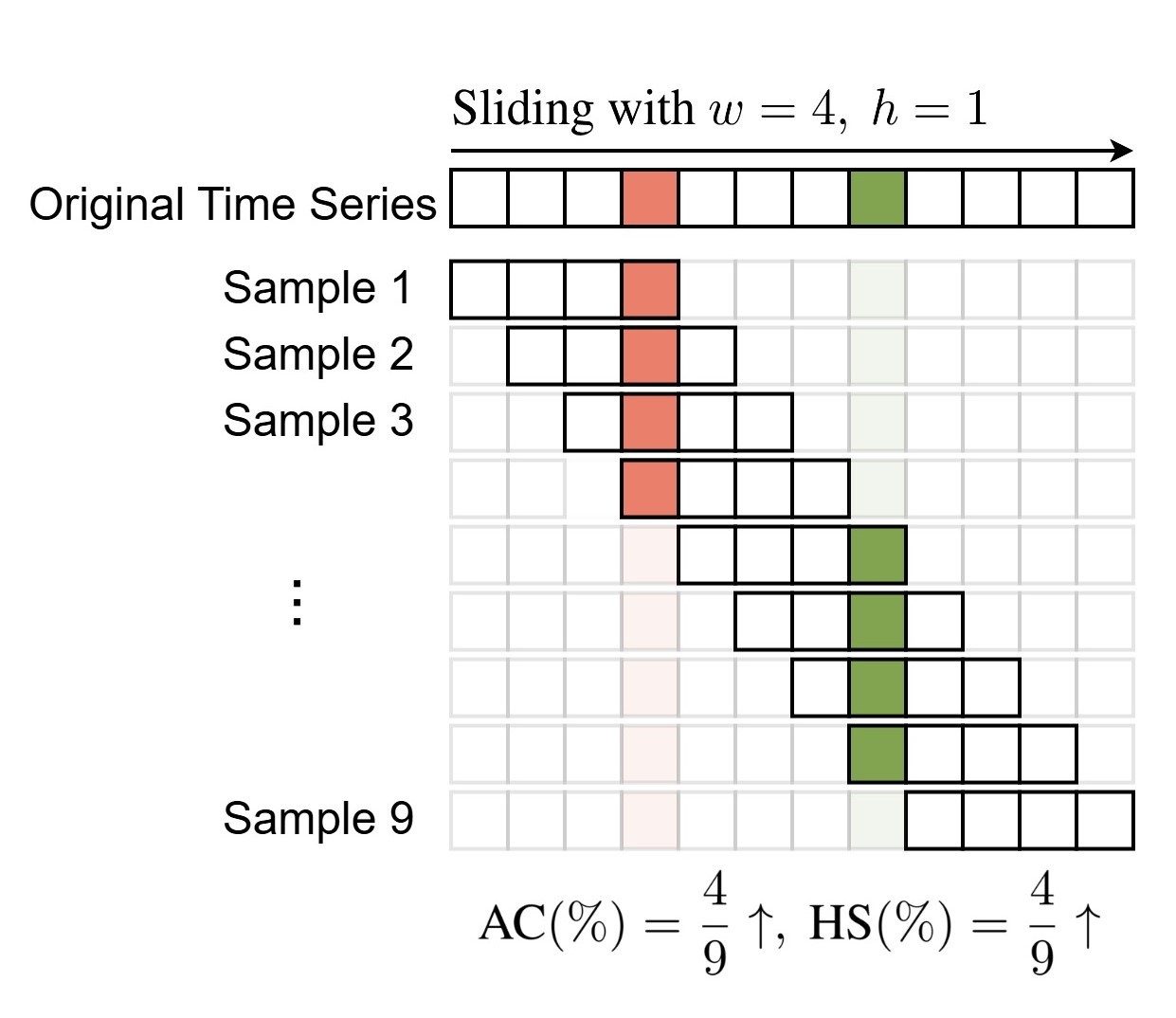}\label{fig:slidinga}}
\hspace{15pt}
\subfloat[no-overlap Sliding Window]{\includegraphics[height=4.5cm]{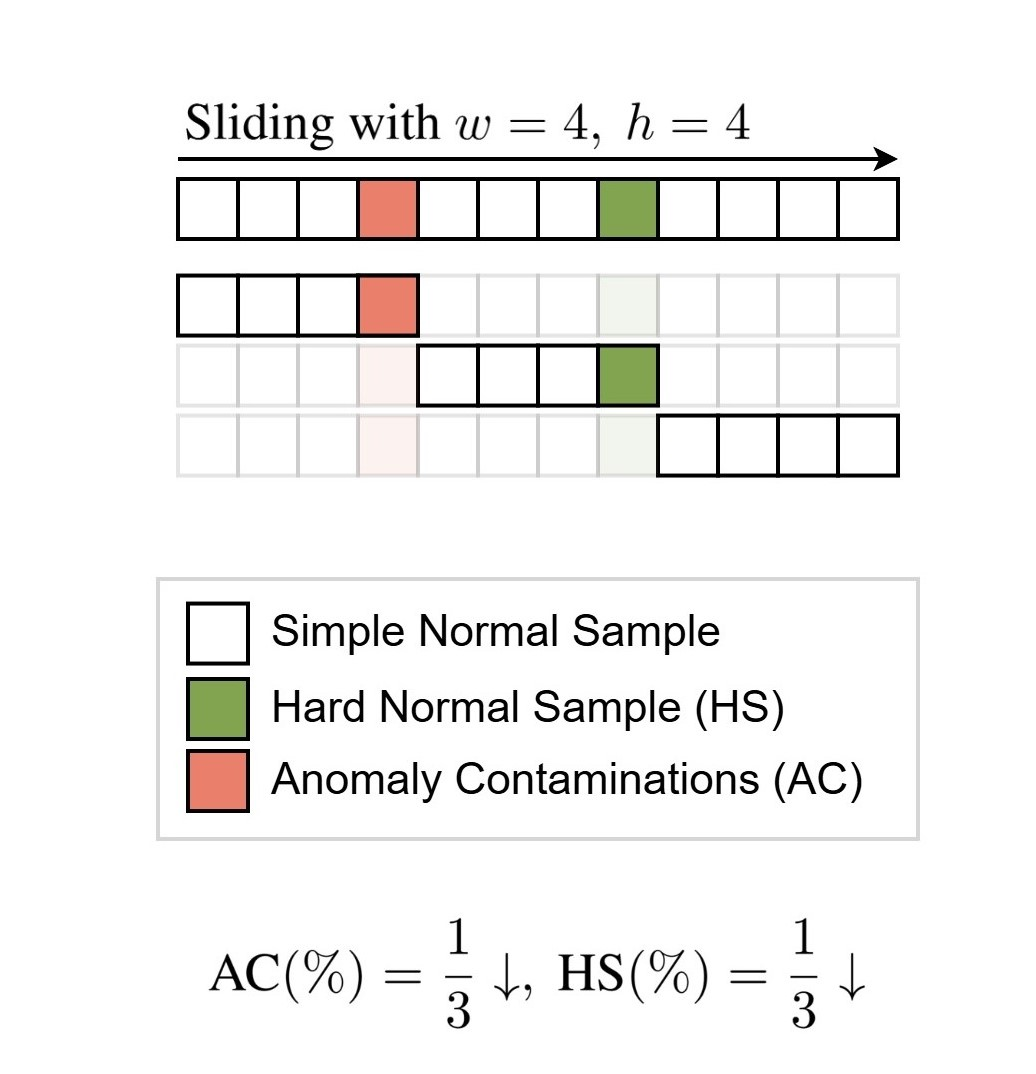}\label{fig:slidingb}}
\hspace{15pt}
\subfloat[Adaptive Sliding Window]{\includegraphics[height=4.5cm]{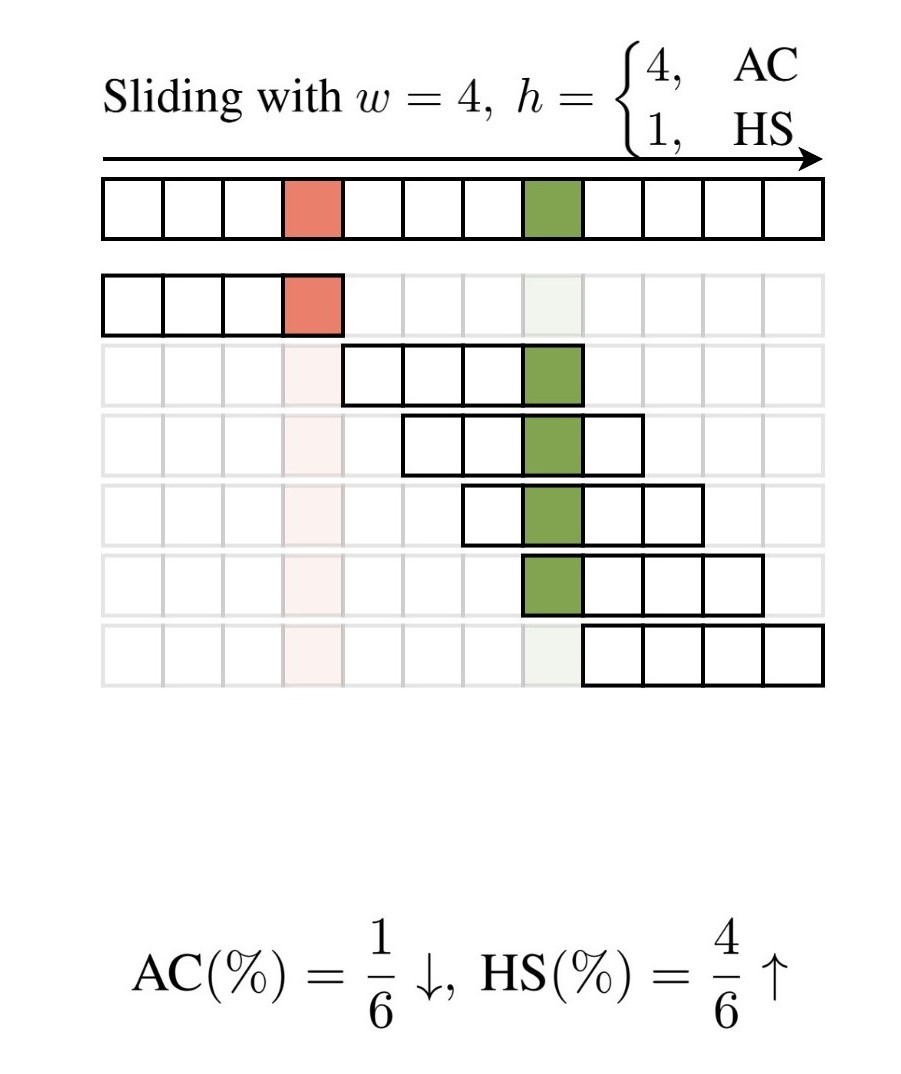}\label{fig:slidingc}}
\caption{
Illustration of sliding window with $w$ as the window size and $h$ as the stride. The sliding window converts the original time series into a series of samples. Ideally, we aim to reduce AC and enrich HS. However, the current methods either (a)~enrich both AC and HS or (b)~reduce them simultaneously. On the contrary, as shown in (c), our adaptive method ensures AC reduction and HS enrichment.
} 
\label{fig:sliding}
\end{figure*}

We leverage the continuous nature of time series to implement an adaptive sliding window module for data augmentation. As illustrated in \cref{fig:slidinga,fig:slidingb}, the fixed stride method treats all samples equally. We aim to adjust strides adaptively by sample types, ensuring AC reduction and HS enrichment, as depicted in \cref{fig:slidingc}.

This is achieved through a set of actions~$\{a^0, a^1, a^2\}$, supports expansion, preservation, and deletion. 
The deletion removes $\mathbf{s}_t$ from $\mathcal{S}_t$, and the preservation maintains $\mathcal{S}_t$ unchanged. 
We use a nearby re-sliding window to generate new samples for expansion.
In this manner, the sliding strides of different samples are adaptively adjusted according to their types, thereby modifying their proportions in the training set.
Our actions are defined as:
\begin{equation}
\label{eq:expand}
    \scalebox{1}{$
    \mathcal{S}_{t+1}=
    \begin{cases}  
        \mathcal{S}_t\bigcup\{\mathbf{s}_{t}^{-w_1}, \mathbf{s}_{t}^{+w_2}, \mathbf{s}_{t}^{-w_2}, \mathbf{s}_{t}^{+w_1}\}, &a = a^0, \\    
        \mathcal{S}_{t}, &a = a^1, \\
        \mathcal{S}_t-\{\mathbf{s}_t\}, &a=a^2,   
    \end{cases}$}
\end{equation}
in which $w_1+w_2=w$, $w_1$ and $w_2$ are coprime. $\mathbf{s}_t^{-\omega_1}$ indicates a re-sliding window that moves $w_1$ steps forward from the current state $\mathbf{s}_{t}$, similarly for other terms. \cref{app:exp} confirms that our design can expand any sample into the training set via a finite amount of actions.

\subsubsection{State Transition}
A state transition function $G$ is introduced to select the next state $\mathbf{s}_{t+1}\in\mathcal{S}_{t+1}$ for investigation. 
$G$ executes two sub-functions, $G_r$ and $G_a$, at probabilities $p$ and $1-p$, to balance exploration and exploitation. 
$G_r$ samples randomly across all states for exploring more samples, while $G_a$ favours AC and HS to improve important samples exploitation.
With a Euclidean distance $d$ to assess the similarity, the state transition function is defined as:
\begin{equation}  \label{eq:sample}
    \scalebox{1}{$
    \mathbf{s}_{t+1}=G_a(\mathcal{S}_{t+1},\mathbf{s}_t,a_t)=
    \begin{cases}  
        \mathop{\arg\min}\limits_{\mathbf{s}\in\mathcal{S}_{t+1}} d(\mathbf{s}_{t},\mathbf{s}), &a_t = a^0, \\  
        \mathop{\arg\max}\limits_{\mathbf{s}\in\mathcal{S}_{t+1}} d(\mathbf{s}_{t},\mathbf{s}), &a_t=a^1,  \\
        \mathop{\arg\min}\limits_{\mathbf{s}\in\mathcal{S}_{t+1}} d(\mathbf{s}_{t},\mathbf{s}), &a_t=a^2.
    \end{cases}$}
\end{equation}
Specifically, if the agent identifies the current state $\mathbf{s}_t$ as AC or HS (i.e., $a_t=a^0$ or $a_t=a^2$), $G_a$ chooses the most similar state to $\mathbf{s}_t$; otherwise, it picks the furthest potential AC or HS, thus boosting the chances of exploring important samples.

\subsubsection{Data Investigation}
We introduce a dual-dimensional reward to assess sample types through parameter and loss behavior. Based on small-loss trick~\cite{arazo2019unsupervised,li2019gradient}, we define loss behavior reward as:
\begin{equation}
    r^l=L(\mathbf{s},\theta).
\end{equation}

The parameter behavior reward is defined by the distance to the parameter behavior center $\Bar{\mathbf{p}}$, as abnormal samples are typically more scattered:
\begin{equation}
    r^p=\Vert \mathbf{p}-\Bar{\mathbf{p}}\Vert_{2}.
\end{equation}


The loss behavior reward $r^l$ indicates the learning difficulty, with AC and HS receiving larger rewards than simple normal samples. The parameter behavior reward $r^p$ reflects the parameter sensitivity, being smaller for normal samples and larger for AC. A dual-dimensional reward $r=R_{\alpha}(r^l,r^p)$ combines them to offer a comprehensive reward for varying samples.
\begin{equation}  \label{eq:reward}
    R_{\alpha}(r^l,r^p)=
    \begin{cases}  
        \alpha r^l+(1-\alpha)(1-r^p), &a = a^0, \\     
        \alpha(1-r^l)+(1-\alpha)(1-r^p), &a = a^1, \\   
        \alpha r^l+(1-\alpha)r^p, &a=a^2,        
    \end{cases}
\end{equation}
in which $\alpha\in[0,1]$ is a factor to balance parameter and loss behaviors.
Both $r^l$ and $r^p$ are normalized. 
Thus, simple normal samples, HS, and AC have different reward tendencies under different actions, enabling effective discriminating.

\subsection{PLDA in Time Series Anomaly Detection}



PLDA operates during the training phase of the TSAD model, augmenting the training set as illustrated in \cref{fig:frameworkb}. Given a time series dataset $\mathcal{X}$, we begin by applying a no-overlap sliding window module to construct the initial training set $\mathcal{S}_0$, which serves as the basis for initial model training. During the $i$-th epoch, PLDA utilizes the behaviors from the optimized TSAD model to augment the current training set, denoted as $\mathcal{S}_i$. The augmented training set, represented as $\mathcal{S}_i^*$, is then redeployed into the TSAD model for additional training. This process is repeated until it reaches the return condition. Following these epochs, the TSAD model finalizes its training on the augmented training set, facilitating anomaly detection.

\subsection{Algorithms Details}


\cref{alg:all} outlines the operation of PLDA within the TSAD model training process. In Step~(3), the time series dataset is initially processed into a no-overlap training set. During the $i$-th training epoch, model training and data augmentation are alternated in Step~(4-7). Specifically, in Step~(5), the TSAD model is trained using the training set. Subsequently, in Step~(6), PLDA is employed for data augmentation by leveraging the parameter behavior and loss behavior recorded during the former training process. After $e$ rounds of alternating iterations, as Steps~(9–11) described, the final augmented training set $\mathcal{S}^*$ is obtained. The model is then repeatedly trained with $\mathcal{S}^*$ until the return condition is reached. This process results in the fully trained TSAD model.


The specific data augmentation process of PLDA is detailed in \cref{alg:PLDA}. 
If this is the first round of training for the TSAD model (i.e., $i=0$), PLDA begins with a warm start.
It leverages stochastic sampling and decision-making to populate its memory, as delineated in Steps~(3-11)
After initializing the training set and the current state in Steps~(12-14), PLDA iteratively enriches the dataset in Steps~(15-26). In this phase, the data investigation module calculates the dual-dimensional reward for the current state in Step~(16). The agent's policy network then determines the action value, selecting the optimal action in Step~(17). The data augmentation module enhances the training set with this action in Step~(18). The subsequent state is then chosen via sampling, and information is saved into memory during Steps~(19-20). The policy and target networks are updated in Steps~(21-25) by extracting mini-batches from memory at each epoch to refine the target network. The policy network experiences an update every $q$ epochs using insights from the target network. After $n$ iterations, this process produces a robustly augmented training set $\mathcal{S}^*$, ready for anomaly detection training.

\begin{algorithm}[tb]
	\caption{PLDA in Time Series Anomaly Detection Training}
	\label{alg:all}
	\begin{algorithmic}[1]
		\STATE {\bfseries Input:} $\mathcal{X}$ - A time series dataset.
		\STATE {\bfseries Output:} 
		$\mathcal{M}_\theta$ - The trained TSAD model.
		\STATE Execute no-overlap sliding window on $\mathcal{X}$ to get the initial training set $\mathcal{S}_0$.
		\FOR{$i=0$ {\bfseries to} $e$}
		\STATE Train $M$ on $\mathcal{S}_i$
		\STATE Augment $\mathcal{S}_i$ to $\mathcal{S}_i^*$ with PLDA
		\ENDFOR
		\STATE $\mathcal{S}^* \leftarrow \mathcal{S}_e^*$ 
		\REPEAT
		\STATE Train $M$ on $\mathcal{S}^*$
		\UNTIL Meet the early stop condition or reach the maximum number of training
		\RETURN $M$
	\end{algorithmic}
\end{algorithm}

\begin{algorithm}[tb]
	\caption{The Process of PLDA}
	\label{alg:PLDA}
	\begin{algorithmic}[1]
		\STATE {\bfseries Input:} $\mathcal{S}_i$ - The original training set.
		\STATE {\bfseries Output:} 
		$\mathcal{S}_i^*$ - The augmented training set.
		\IF{$i==0$} 
		\REPEAT
		\STATE Random sample a $\mathbf{s}$ from $\mathcal{S}_i$ as the current state.
		\STATE Calculate the reward $r$ according to \cref{eq:reward}
		\STATE Random select an action $a$
		\STATE $\mathbf{s}'\leftarrow G(\mathcal{S}_i,\mathbf{s},a)$
		\STATE Push $(\mathbf{s}, r, a, \mathbf{s}')$ into the memory
		\UNTIL Reach the maximum of epoch of warm start
		\ENDIF
		\STATE Initialize $\mathcal{S}_{i, 0}\leftarrow\mathcal{S}_i$
		\STATE Initialize $\mathbf{s}_0\in\mathcal{S}_{i,0}$ randomly
		\STATE Calculate the key parameters.
		\FOR{$t=0$ {\bfseries to} $n$}
		\STATE Calculate the reward $r_t$ according to \cref{eq:reward}
		\STATE $a_t\leftarrow \mathop{\arg\max}\limits_{a}Q_\varphi(\mathbf{s}_t)$
		\STATE Execute $a_t$ on $\mathcal{S}_{i,t}$ to obtain the augmented training set $\mathcal{S}_{i,t+1}$
		\STATE $s_{t+1}\leftarrow G(\mathcal{S}_{i,t},s_t,a_t)$
		\STATE Push $(s_t, r_t, a_t, s_{t+1})$ into the memory
		\REPEAT
		\STATE Sample a mini-batch from the memory
		\STATE Update $Q_{\varphi'}$ according to \cref{eq:varphi}.
		\STATE Update $Q_{\varphi}$ according to $Q_{\varphi'}$ per $q$ epoch 
		\UNTIL Reach maximum number of mini-batches
		\ENDFOR
		\STATE $\mathcal{S}_i^* \leftarrow \mathcal{S}_{i,n}$
		\RETURN $\mathcal{S}_i^*$
	\end{algorithmic}
\end{algorithm}

\section{Experiments}\label{sec:Experiment}

\subsection{Experimental Setup}
\subsubsection{Datasets}
We employ eight publicly available benchmarks to evaluate PLDA: (1) $\textbf{ASD}$ is collected from a large International company. It contains the application server data every 5 minutes in 45 days, in which the first 30 days are normal actions~\cite{li2021multivariate}. (2) $\textbf{MSL}$ is collected by NASA, which shows the condition of the sensor and actuator data from the Mars rover~\cite{hundman2018detecting}. (3) $\textbf{SMAP}$ is also collected by NASA and presents the soil samples and telemetry information used by the Mars rover~\cite{hundman2018detecting}. (4) $\textbf{SMD}$ is collected from a large International company, which contains the server data of every minute in 5 weeks~\cite{su2019robust}. (5) $\textbf{SWaT}$ is collected from an industrial water treatment plant. It contains samples every second in 11 days, in which the first seven days are normal actions\cite{goh2017dataset}. (6) $\textbf{PUMP}$ is collected from a water pump system of a small town~\cite{feng2021time}. (7) $\textbf{DSADS}$ is the motion sensor data collected from 19 activities of daily living and physical activity in 8 subjects~\cite{altun2010comparative}. (8) $\textbf{UCR}$ contains 250 sub-datasets from various natural sources~\cite{dau2019ucr}. We choose Fault, Gait, and Heart as the benchmarks. They are collected from the real-world system, which is more practical and may be contaminated. The statistical details are summarized in \cref{tab:datasets}.

\cref{tab:datasets} provides details and statistics of the datasets we used, such as training, validation, and testing set sizes, dimensionality, and anomaly rates. These multivariate datasets, widely cited in the literature \cite{xu2021anomaly,wu2022timesnet,yang2023dcdetector}, are concatenated for our experiments according to the work~\cite{xu2021anomaly}. We experiment with each UCR entity separately and report average values due to the entity differences. All datasets are officially divided into a training set and a testing set. 


\begin{table}[tb]
	\caption{Detail of benchmark datasets. Dim (Dimensionality) indicates the number of features. AR (anomaly ratio) represents the abnormal proportion of the testing set. The statistical information of UCR datasets are the average value over their sub-datasets. }\label{tab:datasets}
	\centering
	\scalebox{1}{
		\begin{tabular}{lllllr}
			\toprule[1.5pt]
			\multirow{2}{*}{\textbf{Data}} & \multirow{2}{*}{$\bm\#$\textbf{Train}} & \multirow{2}{*}{$\bm\#$\textbf{Valid}} & \multirow{2}{*}{$\bm\#$\textbf{Test}} & \multirow{2}{*}{\textbf{Dim}} & \multirow{2}{*}{\textbf{AR}} \\
			& & & & & \\
			\midrule
			ASD  & 81865  & 20466  & 51840 & 19 & 4.6\%   \\ 
			MSL & 46654  & 11663  & 73729 & 55 & 10.5\%   \\ 
			SMAP & 108146  & 27036  & 427617 & 25 & 12.8\%   \\ 
			SMD & 566724  & 141681  & 708420 & 38 & 4.2\%   \\ 
			SWaT & 380160  & 95040  & 449919 & 51 & 12.1\%   \\ 
			PUMP & 61520  & 15380  & 143401 & 43 & 10.0\%   \\ 
			DSADS & 460800  & 115200  & 564000 & 45 & 31.9\%   \\ 
			Fault & 16106  & 4027  & 43866 & 1 & 1.2\%   \\ 
			Gait & 17733  & 4433  & 42833 & 1 & 0.9\%   \\ 
			Heart & 14800  & 3700  & 40000 & 1 & 0.9\%   \\ 
			\bottomrule[1.5pt]
	\end{tabular}}
\end{table}


%

%

\subsubsection{Benchmark Unsupervised Deep TSAD Models}
Four unsupervised TSAD models are chosen as the benchmarks: (1) \textbf{TcnED} is a temporal anomaly detection algorithm based on temporal convolutional networks, capturing temporal dependencies through causal convolution~\cite{bai2018empirical}. (2) \textbf{TranAD} is an anomaly detection and diagnosis model that leverages a deep transformer network and employs an attention-based sequence encoder~\cite{tuli2022tranad}. (3) \textbf{NeuTral} proposes a paradigm for data transformation targeting time series and tabular data to enable anomaly detection~\cite{qiu2021neural}. (4) \textbf{NCAD} is a self-supervised anomaly detection algorithm that identifies boundaries by injecting synthetic anomalies into existing data for learning~\cite{ijcai2022p394}. 
These four models have varied network structures, verifying the generality of PLDA. Besides, they have relatively small parameter sizes, making them more suitable for PLDA.

\subsubsection{Competing Data Augmentation Methods}
Three methods have been selected as competitors, and we refer to them as ORIG, PI, and LOSS, based on their operating principles. Detailed explanations are provided as follows: (1) \textbf{ORIG} utilizes the original training set without augmentation. (2) \textbf{PI} assesses the importance of model parameters and updates only the critical parameters during each epoch~\cite{xia2020robust}. (3) \textbf{LOSS} identifies AC by examining high loss values or rapid changes in loss, filtering out such AC using a predefined threshold~\cite{li2022robust}.

\subsubsection{Parameter Settings and Implementations}
All the competitor models utilize their default hyper-parameter settings. For PLDA, the recommended hyperparameters are provided in the sensitivity test. Given that our PLDA approach uses only a small amount of data, there is a risk that the original model might overfit under the same parameter configurations. To ensure a fair comparison across different models, we implemented an early stopping mechanism and limited the experiment to a maximum of 50 epochs.

Our experiments are conducted on a workstation with an Intel Xeon Silver 4210R CPU, a single NVIDIA TITAN RTX GPU, and 64 GB of RAM. All the TSAD models used in our experiments are implemented in Python. The four TSAD benchmark models (TcnED, TranAD, NeuTral, NCAD) are implemented by the open-source project DeepOD\footnote{\url{https://github.com/xuhongzuo/deepod}}, and three data augmentation competitors (ORIG, PI, LOSS) are implemented according to their paper. 


\subsubsection{Evaluation Protocol}
We adopt the best F1-score~\cite{zhang2021unsupervised,xu2022calibrated} to evaluate the performance, which is determined by searching all possible anomaly thresholds. Besides, we employ the commonly-used adjustment technique for a fair comparison~\cite{li2021multivariate}: for each ground-truth anomaly segment, we use the maximum score as the score of all points in that segment.

\subsection{Effectiveness in Real-World Datasets}
We evaluate the effectiveness of our PLDA in enhancing detection performance. Four TSAD models, enhanced by four data augmentation methods, are evaluated across ten datasets.
Each experimental set is repeated five times for reliability, with standard deviation calculated.

\cref{tab:DAeff} organizes the outcomes into four sub-tables, with each sub-table dedicated to a specific TSAD model. These sub-tables facilitate comparison between the baseline model and the data augmentation methods. The concluding rows emphasize the average F1-score and the improvements over the baseline model. PLDA demonstrates exceptional performance, increasing average F1-scores by 3.88\% to 8.03\% and enhancing detection capabilities in 7 to 9 datasets. Note that PLDA shows less prominent performance on datasets such as MSL, SMAP, and SWaT. In addition to our methods, competitors like PI and LOSS also display only minor improvements on these datasets, indicating a common phenomenon. This likely occurs because these datasets contain fewer native AC, limiting PLDA's ability to improve performance by reducing AC. Moreover, NeuTral shows significant improvement on the MSL dataset, suggesting that the enrichment of HS also contributes to enhanced performance.
This proves that correctly distinguishing between these data types helps establish clearer decision boundaries and enhances overall performance.


\begin{table*}[!t]
	\setlength{\tabcolsep}{1.2pt}
	\renewcommand{\arraystretch}{1.5}
	\caption{F1-score $\pm$ standard deviation on ten datasets. Four data augmentation methods on four deep TSAD models are tested. All the results are in \%, and the best ones are in \textbf{bold}.}\label{tab:DAeff}
	\centering
	\scalebox{0.94}{
		\begin{tabular}{l|cccc|cccc|cccc|cccc}
			\bottomrule[1.5pt]
			& \multicolumn{4}{c|}{\textbf{TcnED}} & \multicolumn{4}{c|}{\textbf{TranAD}} & \multicolumn{4}{c|}{\textbf{NeuTral}} & \multicolumn{4}{c}{\textbf{NCAD}} \\\cline{2-17} 
			\textbf{Data} & ORIG  & PI & LOSS & PLDA  & ORIG  & PI & LOSS  & PLDA  & ORIG  & PI & LOSS & PLDA  & ORIG  & PI & LOSS & PLDA  \\
			
			\hline
			ASD & $\text{76.5}_{\pm\text{3.4}}$ & $\text{78.8}_{\pm\text{3.0}}$ & $\text{78.5}_{\pm\text{2.2}}$ &  $\text{\textbf{85.2}}_{\bm\pm\text{\textbf{0.2}}}$ 
			
			& $\text{65.5}_{\pm\text{4.4}}$ & $\text{68.4}_{\pm\text{3.2}}$  & $\text{66.6}_{\pm\text{3.6}}$ & 
			$\text{\textbf{85.0}}_{\bm\pm\text{\textbf{0.7}}}$ 
			
			& $\text{61.4}_{\pm\text{2.2}}$ &  $\text{62.6}_{\pm\text{1.8}}$ 
			& $\text{60.1}_{\pm\text{1.5}}$ &$\text{\textbf{63.6}}_{\bm\pm\text{\textbf{0.6}}}$    
			
			& $\text{73.4}_{\pm\text{1.9}}$ & $\text{73.6}_{\pm\text{1.5}}$    & $\text{75.6}_{\pm\text{1.9}}$     & $\text{\textbf{77.2}}_{\bm\pm\text{\textbf{3.4}}}$\\
			
			MSL & $\text{89.2}_{\pm\text{0.3}}$      & $\text{88.6}_{\pm\text{1.0}}$ &  $\text{\textbf{89.3}}_{\bm\pm\text{\textbf{0.5}}}$  & $\text{89.0}_{\pm\text{1.1}}$ 
			
			& $\text{75.6}_{\pm\text{0.9}}$  &  $\text{75.6}_{\pm \text{1.6}}$  & $\text{71.9}_{\pm\text{0.6}}$  &
			$\text{\textbf{76.7}}_{\bm\pm\text{\textbf{2.4}}}$ 
			
			& $\text{72.2}_{\pm\text{4.3}}$ & $\text{75.4}_{\pm\text{6.6}}$ 
			&  $\text{72.3}_{\pm\text{4.1}}$ &  $\text{\textbf{81.0}}_{\bm\pm\text{\textbf{1.8}}}$
			
			& $\text{80.4}_{\pm\text{2.0}}$  &  $\text{80.6}_{\pm\text{2.2}}$ &  $\text{81.3}_{\pm\text{1.8}}$    & 
			$\text{\textbf{82.0}}_{\bm\pm\text{\textbf{1.5}}}$    \\
			
			SMAP & $\text{70.7}_{\pm\text{1.0}}$  & $\text{70.2}_{\pm\text{0.5}}$ & $\text{70.6}_{\pm\text{0.8}}$  &  $\text{\textbf{71.1}}_{\bm\pm\text{\textbf{0.2}}}$     
			
			& $\text{\textbf{70.5}}_{\pm\text{\textbf{0.1}}}$  & $\text{70.5}_{\pm\text{0.2}}$ & $\text{70.5}_{\pm\text{0.1}}$ & $\text{70.1}_{\pm\text{0.2}}$ 
			
			& $\text{47.1}_{\pm\text{0.9}}$ &  $\text{47.8}_{\pm\text{0.7}}$ &  $\text{47.6}_{\pm\text{0.5}}$&  $\text{\textbf{49.7}}_{\bm\pm\text{\textbf{0.8}}}$ 
			
			& $\text{69.8}_{\pm\text{0.5}}$ & $\text{69.7}_{\pm\text{0.6}}$ &  $\text{69.9}_{\pm\text{0.4}}$&  $\text{\textbf{70.0}}_{\bm\pm\text{\textbf{0.5}}}$ \\
			
			SMD & $\text{80.9}_{\pm\text{0.8}}$  & $\text{79.4}_{\pm\text{0.8}}$& $\text{81.2}_{\pm\text{0.8}}$ &  $\text{\textbf{81.6}}_{\bm\pm\text{\textbf{1.3}}}$ 
			
			& $\text{80.4}_{\pm\text{1.6}}$ &   $\text{80.6}_{\pm\text{3.2}}$&  $\text{\textbf{82.2}}_{\bm\pm\text{\textbf{2.4}}}$ & $\text{79.9}_{\pm\text{1.5}}$ 
			
			&$\text{59.8}_{\pm\text{2.5}}$ & $\text{\textbf{61.6}}_{\bm\pm\text{\textbf{3.4}}}$ & $\text{57.4}_{\pm\text{2.2}}$ & $\text{55.8}_{\pm\text{2.5}}$     & 
			
			$\text{76.4}_{\pm\text{1.0}}$  &  $\text{76.4}_{\pm\text{1.0}}$      & $\text{76.8}_{\pm\text{2.1}}$     &  $\text{\textbf{79.3}}_{\bm\pm\text{\textbf{4.5}}}$   \\
			
			SWaT & $\text{\textbf{84.3}}_{\bm\pm\text{\textbf{0.8}}}$ &$\text{84.2}_{\pm\text{0.3}}$ &$\text{84.2}_{\pm\text{0.1}}$  & $\text{84.1}_{\pm\text{0.2}}$ & 
			
			$\text{84.3}_{\pm\text{0.3}}$  & $\text{84.0}_{\pm\text{0.3}}$ & $\text{84.3}_{\pm\text{0.4}}$  & $\text{\textbf{84.3}}_{\bm\pm\text{\textbf{0.3}}}$      
			
			& $\text{79.0}_{\pm\text{1.9}}$       &  $\text{82.3}_{\pm\text{0.3}}$ &  $\text{82.1}_{\pm\text{0.4}}$ &  $\text{\textbf{82.7}}_{\bm\pm\text{\textbf{0.8}}}$ 
			
			& $\text{94.7}_{\pm\text{0.4}}$  & $\text{\textbf{94.8}}_{\bm\pm\text{\textbf{0.4}}}$     & $\text{94.4}_{\pm\text{0.9}}$ & $\text{94.7}_{\pm\text{0.7}}$        \\
			
			PUMP & $\text{74.8}_{\pm\text{15.9}}$ & $\text{84.9}_{\pm\text{3.3}}$ &  $\text{88.1}_{\pm\text{1.2}}$ &  $\text{\textbf{90.5}}_{\bm\pm\text{\textbf{0.3}}}$
			
			& $\text{70.4}_{\pm\text{13.6}}$ &  $\text{87.8}_{\pm\text{0.2}}$&  $\text{84.9}_{\pm\text{5.6}}$ &  $\text{\textbf{88.3}}_{\bm\pm\text{\textbf{0.5}}}$ 
			
			& $\text{58.3}_{\pm\text{7.3}}$ & $\text{55.0}_{\pm\text{1.0}}$ & $\text{54.2}_{\pm\text{0.6}}$ & $\text{\textbf{63.9}}_{\bm\pm\text{\textbf{7.7}}}$ 
			
			& $\text{97.0}_{\pm\text{0.5}}$  &  $\text{97.3}_{\pm\text{0.4}}$ & $\text{96.1}_{\pm\text{0.6}}$       &  $\text{\textbf{97.2}}_{\bm\pm\text{\textbf{0.5}}}$ \\
			
			DSADS & $\text{95.5}_{\pm\text{0.2}}$  &  $\text{96.1}_{\pm\text{1.1}}$ &  $\text{96.2}_{\pm\text{0.6}}$ &  $\text{\textbf{97.8}}_{\bm\pm\text{\textbf{0.4}}}$
			
			& $\text{95.7}_{\pm\text{0.6}}$  & $\text{94.5}_{\pm\text{0.9}}$  & $\text{95.0}_{\pm\text{0.3}}$  &  $\text{\textbf{96.9}}_{\bm\pm\text{\textbf{0.2}}}$ 
			
			& $\text{96.1}_{\pm\text{0.3}}$  &  $\text{\textbf{96.6}}_{\bm\pm\text{\textbf{0.2}}}$ & $\text{95.4}_{\pm\text{0.9}}$ &  $\text{96.3}_{\pm\text{0.2}}$ 
			
			& $\text{92.9}_{\pm\text{0.2}}$  & $\text{92.8}_{\pm\text{0.5}}$ &  $\text{93.6}_{\pm\text{1.8}}$ &  $\text{\textbf{94.1}}_{\bm\pm\text{\textbf{0.5}}}$ \\
			
			Fault & $\text{95.2}_{\pm\text{7.4}}$  & $\text{93.4}_{\pm\text{8.6}}$ &  $\text{95.8}_{\pm\text{6.0}}$& $\text{\textbf{98.9}}_{\bm\pm\text{\textbf{1.8}}}$     
			
			& $\text{\textbf{43.2}}_{\bm\pm\text{\textbf{24.3}}}$ & $\text{43.0}_{\pm\text{24.3}}$ & $\text{43.1}_{\pm\text{24.3}}$ & $\text{43.0}_{\pm\text{24.2}}$ 
			
			& $\text{79.9}_{\pm\text{10.2}}$ &  $\text{87.5}_{\pm\text{4.9}}$ &  $\text{83.1}_{\pm\text{8.1}}$ &  $\text{\textbf{87.5}}_{\bm\pm\text{\textbf{9.2}}}$ 
			
			& $\text{60.8}_{\pm\text{9.7}}$  &  $\text{68.0}_{\pm\text{15.9}}$ &  $\text{68.1}_{\pm\text{10.1}}$ &  $\text{\textbf{76.9}}_{\bm\pm\text{\textbf{2.1}}}$ \\
			
			Gait  & $\text{95.8}_{\pm\text{4.5}}$ &  $\text{\textbf{97.7}}_{\bm\pm\text{\textbf{2.1}}}$ &  $\text{97.4}_{\pm\text{2.0}}$& $\text{95.4}_{\pm\text{5.5}}$     
			
			& $\text{69.1}_{\pm\text{8.1}}$  & $\text{68.3}_{\pm\text{25.1}}$ & $\text{66.9}_{\pm\text{9.6}}$ &  $\text{\textbf{85.8}}_{\bm\pm\text{\textbf{12.7}}}$  
			
			&$\text{91.0}_{\pm\text{3.0}}$ & $\text{85.2}_{\pm\text{5.4}}$ & $\text{87.7}_{\pm\text{5.5}}$ &  $\text{\textbf{93.5}}_{\bm\pm\text{\textbf{4.4}}}$ 
			
			& $\text{36.9}_{\pm\text{5.8}}$  &  $\text{40.1}_{\pm\text{9.4}}$ &  $\text{44.0}_{\pm\text{9.3}}$  &  $\text{\textbf{65.2}}_{\bm\pm\text{\textbf{13.7}}}$ \\

			Heart& $\text{77.5}_{\pm\text{2.7}}$  & 
			$\text{76.1}_{\pm\text{1.2}}$ &  $\text{78.3}_{\pm\text{3.7}}$&  $\text{\textbf{80.2}}_{\bm\pm\text{\textbf{4.2}}}$
			
			& $\text{75.6}_{\pm\text{0.1}}$  &  $\text{76.2}_{\pm\text{0.7}}$ &  $\text{75.7}_{\pm\text{0.3}}$ &  $\text{\textbf{79.0}}_{\bm\pm\text{\textbf{2.9}}}$ 
			
			& $\text{84.4}_{\pm\text{2.9}}$ &  $\text{84.5}_{\pm\text{1.6}}$ & $\text{84.3}_{\pm\text{4.4}}$ &  $\text{\textbf{85.4}}_{\bm\pm\text{\textbf{2.0}}}$ 
			
			& $\text{73.8}_{\pm\text{3.8}}$  &  $\text{74.0}_{\pm\text{3.6}}$ & $\text{72.0}_{\pm\text{8.1}}$      &  $\text{\textbf{75.5}}_{\bm\pm\text{\textbf{7.2}}}$  \\
			
			\hline
			Avg & $\text{84.0}_{\pm\text{3.7}}$ & $\text{84.9}_{\pm\text{2.2}}$ & $\text{86.0}_{\pm\text{1.8}}$ & $\text{\textbf{87.3}}_{\bm\pm\text{\textbf{1.6}}}$ 
			
			& $\text{73.0}_{\pm\text{5.4}}$  & $\text{74.9}_{\pm\text{5.9}}$ & $\text{74.1}_{\pm\text{4.7}}$ & $\text{\textbf{78.9}}_{\bm\pm\text{\textbf{4.6}}}$ 
			
			& $\text{72.9}_{\pm\text{3.5}}$    & $\text{73.8}_{\pm\text{2.6}}$ &  $\text{72.4}_{\pm\text{2.8}}$ & $\text{\textbf{76.0}}_{\bm\pm\text{\textbf{3.0}}}$ 
			
			& $\text{75.6}_{\pm\text{2.6}}$  & $\text{76.7}_{\pm\text{3.6}}$ & $\text{77.2}_{\pm\text{3.7}}$      & $\text{\textbf{81.2}}_{\bm\pm\text{\textbf{3.5}}}$ \\
			
			Imp(\%) & -  & 1.06 & 2.29 & \textbf{3.88} & -  & 2.57 & 1.47 & \textbf{8.03} & -  & 1.27 & -0.69 & \textbf{4.17}& -  & 1.45 & 2.14  & \textbf{7.36}\\    
			\toprule[1.5pt]
	\end{tabular}}
\end{table*}

\subsection{Robustness in Contaminated Training Set}
We evaluate PLDA's ability to boost TSAD models' robustness to training set contamination.
Four TSAD models, enhanced by four data augmentation methods, are evaluated on the ASD dataset. ASD is chosen since it is an uncontaminated training set, facilitating a more precise evaluation of the impact of contamination.
The training set is randomly injected with anomalies from the testing set with a ratio from 0\% to 20\%.

\begin{figure}
\centering
\includegraphics[width=0.45\textwidth]{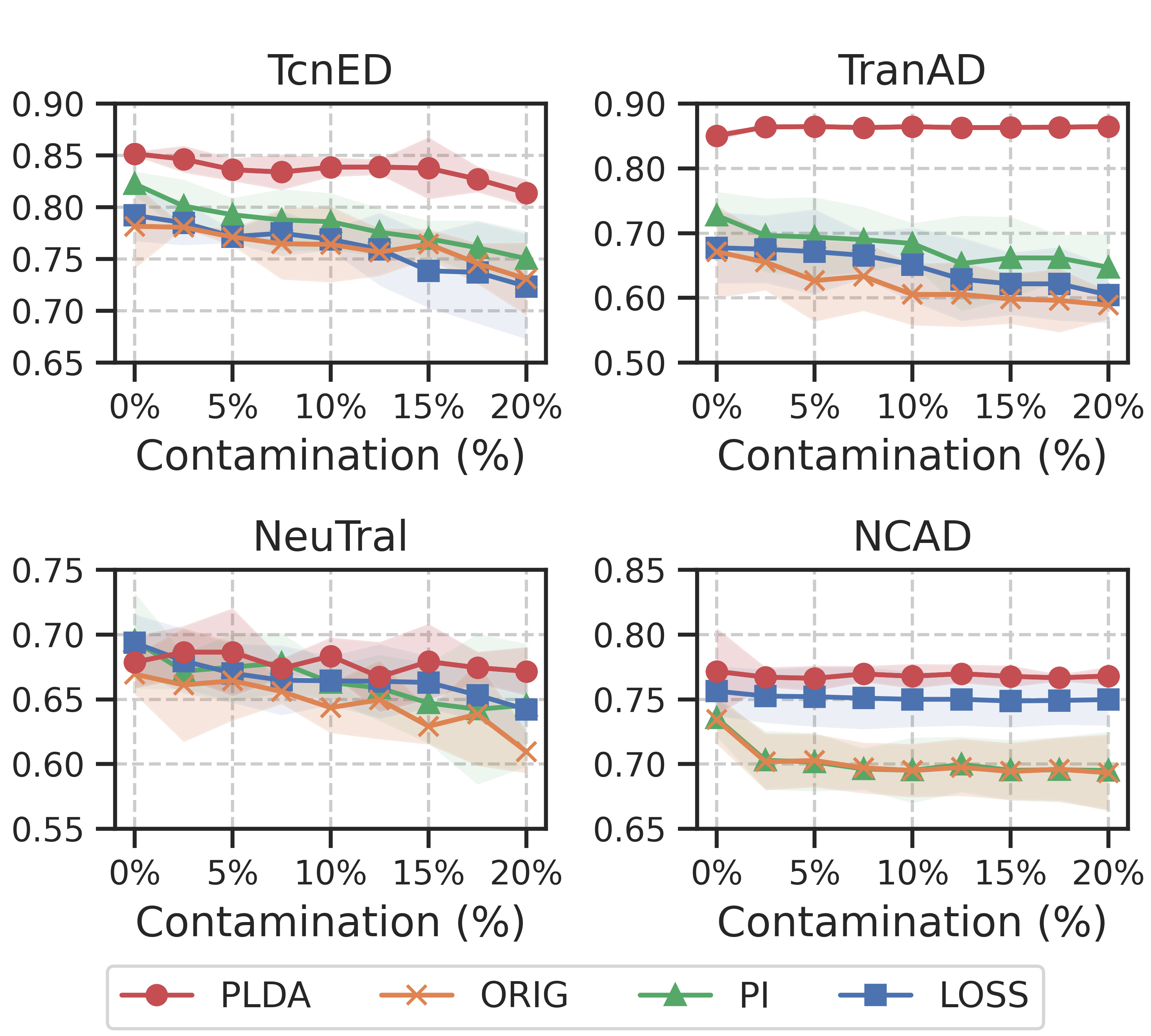}
\caption{F1 results of the four TSAD models augmented by four data augmentation methods on the training set with different contamination rates.} \label{fig:diff_Orate}
\end{figure}

\cref{fig:diff_Orate} shows the average F1-score and its standard deviation over five independent runs. As contamination increases, the original models' performance declines, indicating inadequate robustness. In contrast, PLDA significantly stabilizes the F1-scores, reducing the impact of training set contamination and improving TSAD models' robustness. TcnED, TranAD, and NeuTral, in particular, benefit from PLDA, maintaining high robustness across different contamination levels. NCAD performs consistently across all methods but achieves its best performance with PLDA.
Overall, PLDA served as a protective measure against data contamination, thus enhancing the reliability of anomaly detection.
These findings indicate that PLDA effectively strengthens TSAD models, ensuring stability and accuracy in various conditions.

\subsection{Effectiveness in Addressing AC and HS.}
This experiment evaluates the effectiveness of PLDA in addressing AC and HS.
We track the proportions of AC and HS in the training set when applying PLDA to the TSAD model. 
The labels of HS are determined as follows: We first record the loss value of each data sample during the original training phases, and then we designate normal samples that exceed the loss threshold as HS. In this experiment, our main focus is whether PLDA can distinguish between HS and AC, not its detection performance. Therefore, we simulate the training set using the testing set with ground truth labels. We avoid using a training set with artificially injected AC since the original training set may also contain AC, leading to inaccurate statistics. 
With accurate labels, samples with high loss values but not labeled as Ac are considered HS.

\begin{figure}
\centering
\includegraphics[width=0.33\textwidth]{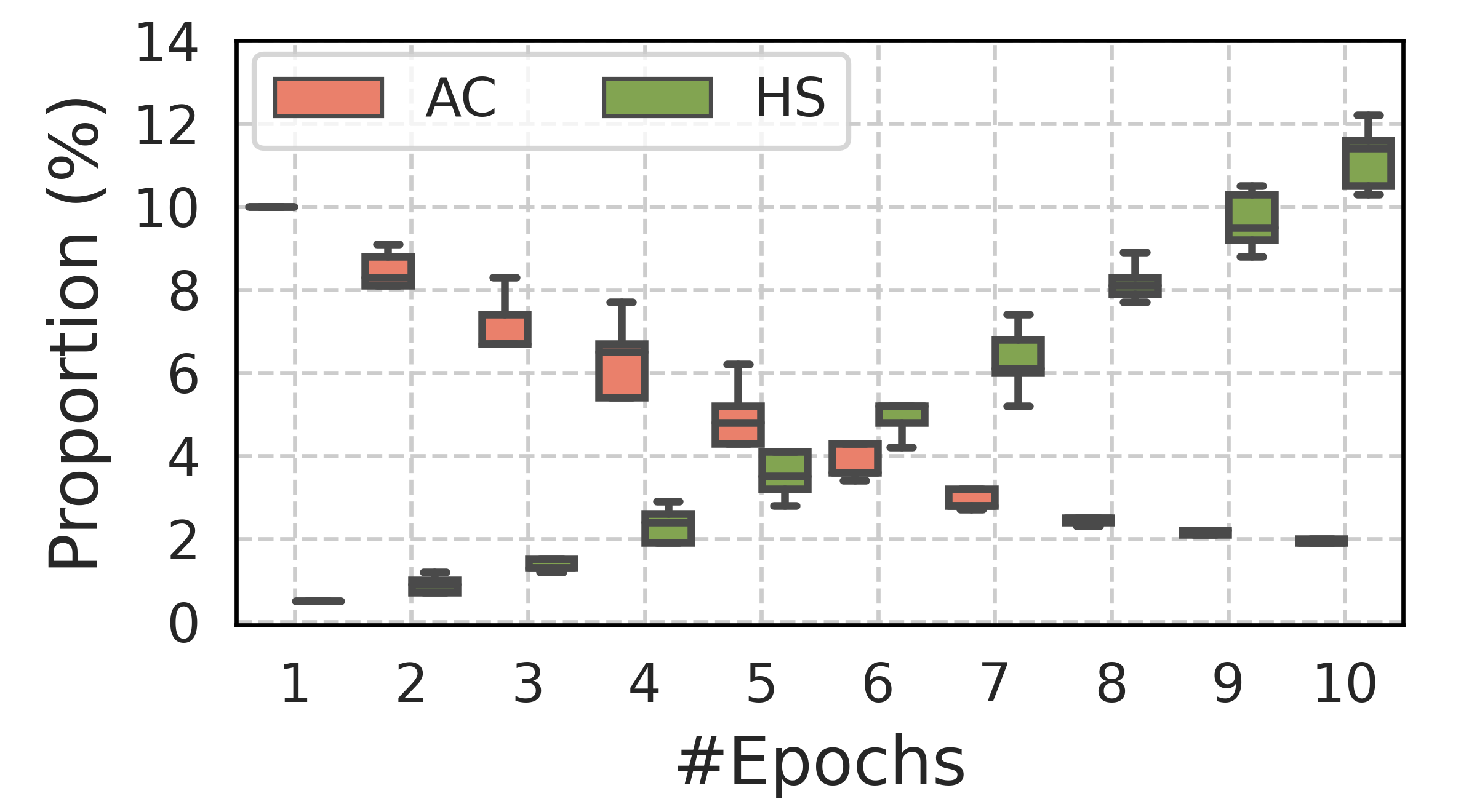}
\caption{Proportion of AC and HS in the training set as epoch increases.} \label{fig:propotion}
\end{figure}

\cref{fig:propotion} provides a comprehensive overview of results obtained from five independent experimental runs. As the training epochs advance, the proportion of AC significantly decreases from 10\% to 2\%, while the proportion of HS increases from 1\% to 11\%. This trend clearly illustrates PLDA's capability to effectively distinguish between AC and HS. Furthermore, it demonstrates PLDA's ability to leverage the adaptive sampling module to adjust data proportions within the training set. These adjustments result in improved performance, even with a reduced dataset size, thereby enhancing the model's reliability and efficiency across various applications.

\subsection{Scalability Test}\label{sec:scalability}

To assess the scalability of PLDA and its competitors, we measure their time efficiency across different dimensions and data sizes.  
We construct synthetic datasets with a constant length of 100,000 across various dimensions (i.e., \{8, 16, 32, 64, 128, 256, 512\}) and datasets with a constant dimension of 16 but increasing lengths (8,000 to 512,000, doubling). 

\begin{figure}
\centering
\includegraphics[width=0.5\textwidth]{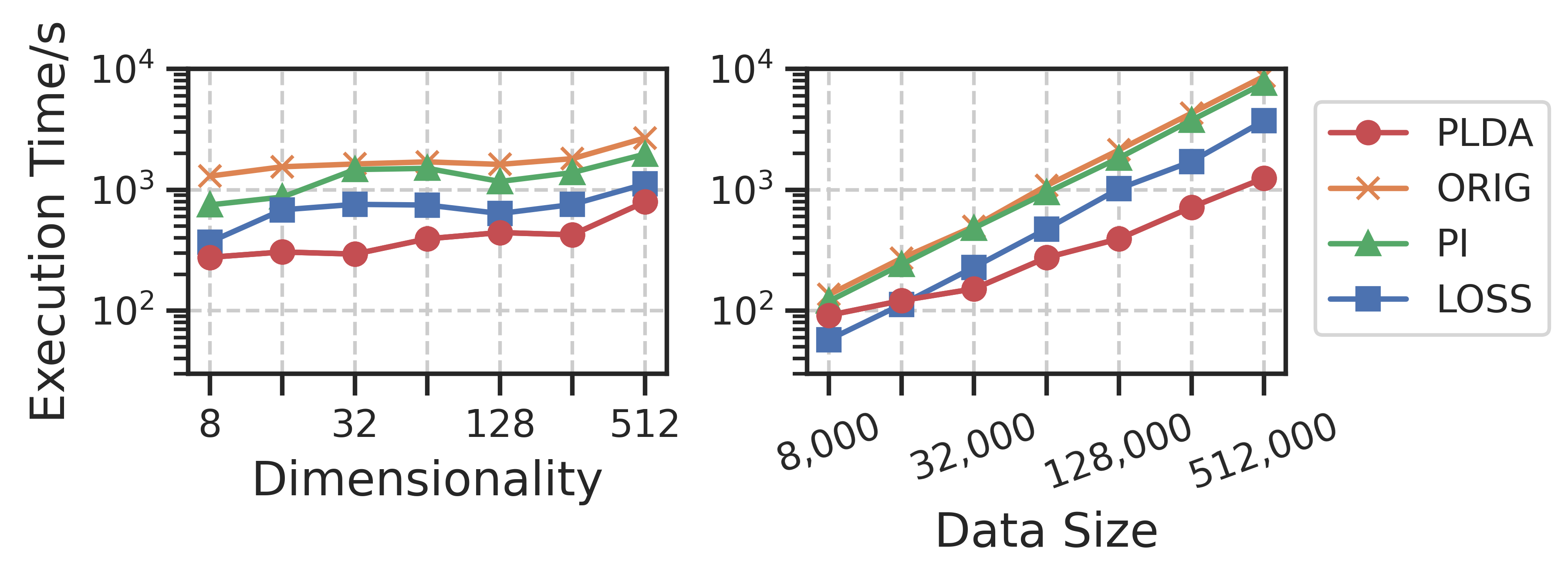}
\caption{Scalability test results w.r.t. dimensionality and data size.} \label{fig:scal}
\end{figure}

\Cref{fig:scal} shows the training time required over 50 epochs. In both conditions, the original model is the least time-efficient. In contrast, all three data augmentation methods save time, with PLDA offering the most significant improvement. However, PLDA's efficiency decreases with smaller data sizes due to the computational overhead of calculating gradient responses for each sample. Despite this, PLDA effectively reduces the training set size using an optimized sliding window module, enhancing overall efficiency. Therefore, while the computational burden of data analysis results in lower time efficiency for smaller datasets, the advantages of PLDA become increasingly evident as the dataset size grows.


\subsection{The Amount of Data Used}\label{sec:dataused}

We introduce a novel adaptive sliding window module that significantly enhances detection performance while requiring only a fraction of the typical training set size. In this study, we conduct a detailed comparison of the training set data usage between the PLDA method and the original dataset.

\cref{tab:dataused} provides a comprehensive summary of the experimental results, including the dataset names, the original training set sizes, the reduced data sizes after implementing PLDA, and the corresponding percentage reductions. Remarkably, PLDA can utilize between 4.4\% and 26.5\% of the original training sets on average, achieving substantial performance improvements. This efficiency illustrates PLDA's capability to expand the training set smartly, effectively reducing AC and augmenting with essential HS through its adaptive approach. Thus, PLDA strengthens anomaly detection robustness while efficiently reducing the required training set size.


\begin{table}
	\setlength{\tabcolsep}{0.6pt}
	\renewcommand{\arraystretch}{1.3}
	\caption{The size of training set used by PLDA on different TSAD models and datasets. Pct (percentage) denotes the proportion of the augmented training set size relative to the original training set size.
	}
	\centering
	\scalebox{1}{
		\label{tab:dataused}
		\begin{tabular}{cc|cc|cc|cc|cc}
			\bottomrule[1.5pt]
			\multirow{2}{*}{\textbf{Data}} & \multirow{2}{*}{$\bm\#$\textbf{Train}} & \multicolumn{2}{c|}{\textbf{TcnED}} & \multicolumn{2}{c|}{\textbf{TranAD}} & \multicolumn{2}{c|}{\textbf{NeuTral}} & \multicolumn{2}{c}{\textbf{NCAD}} \\
			&                           & $\#$Train     & Pct     & $\#$Train     & Pct     & $\#$Train     & Pct     & $\#$Train     & Pct    \\
			\hline
			ASD                       & 81865                      & 13351       & 16.3\%     & 16383        & 20.0\%     & 12470        & 15.2\%      & 2734        & 3.3\%     \\
			MSL                       & 46654                      & 25598       & 54.9\%     & 25743        & 55.2\%     & 1726         & 3.7\%       & 3407        & 7.3\%     \\
			SMAP                      & 108146                     & 5726        & 5.3\%      & 17120        & 15.8\%     & 9508         & 8.8\%       & 5044        & 4.7\%     \\
			SMD                       & 566724                     & 19660       & 3.5\%      & 313363       & 55.3\%     & 18890        & 3.3\%       & 23613       & 4.2\%     \\
			SWaT                      & 380160                     & 19958       & 5.3\%      & 63085        & 16.6\%     & 12672        & 3.3\%       & 17680       & 4.7\%     \\
			PUMP                      & 61520                      & 5789        & 9.4\%      & 2219         & 3.6\%      & 1819         & 3.0\%       & 1536        & 2.5\%     \\
			DSADS                    & 460800                     & 39388       & 8.6\%      & 18359        & 4.0\%      & 6900         & 1.5\%       & 2050        & 0.4\%     \\
			Fault                     & 16106                      & 6713        & 41.7\%     & 6549         & 40.7\%     & 5309         & 33.0\%      & 723         & 4.5\%     \\
			Gait                      & 17733                      & 3513        & 19.8\%     & 8625         & 48.6\%     & 4520         & 25.5\%      & 1272        & 7.2\%     \\
			Heart               & 14800                      & 2062        & 13.9\%     & 734          & 5.0\%      & 1025         & 6.9\%       & 751         & 5.1\%     \\
			\hline
			Avg                   & 175451                     & 14176       & 17.9\%     & 47218        & 26.5\%     & 7484         & 10.4\%      & 5881        & 4.4\%   \\ 
			\toprule[1.5pt]
	\end{tabular}}
\end{table}

\subsection{Sensitivity Test}\label{app:sensitivity}

This experiment investigates the sensitivity of PLDA to three key hyperparameters: $e$ (the number of data augmentation epochs), $a$ (the balance weight between parameter behavior and loss behavior), and $k$ (the number of key parameters) for optimal parameterization in practical use.
Adjusting one hyper-parameter while fixing others, we explore $e$ from $1$ to $32$, $a$ through $\{0.1, 0.3, 0.5, 0.7, 0.9\}$, and $k$ from $200$ to $128,000$.

\begin{figure}
    \centering
    \includegraphics[width=0.48\textwidth]{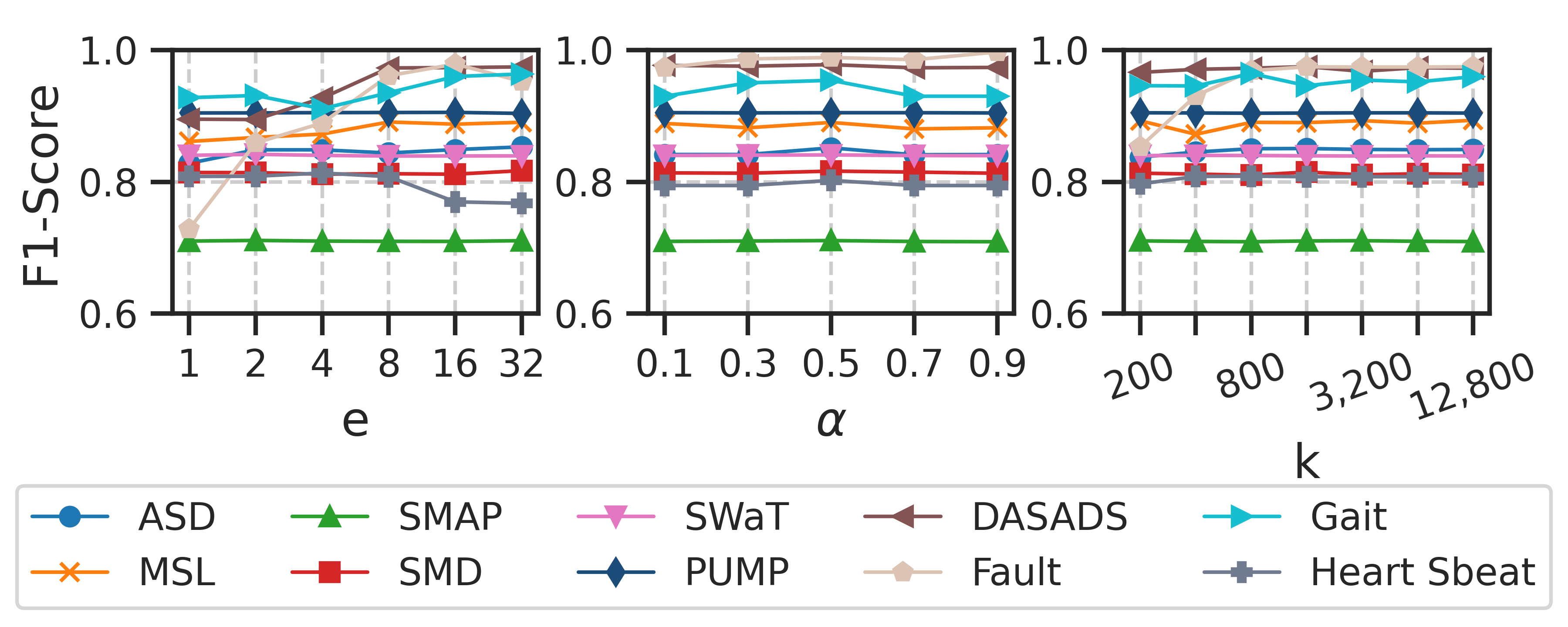}
    \caption{F1-scores of PLDA with different hyper-parameter settings.}
    \label{fig:sensitivity}
\end{figure}

\cref{fig:sensitivity} illustrates the F1-score of PLDA under various parameter settings across ten datasets. 
As $e$ increases, PLDA has more opportunities to augment the training set, leading to an upward trend in the F1-score. However, when $e$ exceeds $8$, this trend diminishes, and some datasets show a slight decline. This may be due to the fixed window size of $30$, which allows PLDA to expand any sub-sequence in the training set up to $8$ times. Beyond this, the incremental impact of up-sampling HS on detection performance decreases. Considering the computational overhead, PLDA sets $e=10$ for data augmentation.
$\alpha$ balances the weight between parameter and loss behavior rewards, with a larger $\alpha$ increasing the emphasis on loss reward. Experiments show it has little effect on the F1-score, likely because both parameter and loss behavior rewards capture purpose characteristics effectively, significantly impacting the results. In this paper, $\alpha$ is set to $0.5$ for practical use.
Overall, $k$ positively influences the F1-score. This is because $k$ indicates the level of detail in behavioral analysis—the larger $k$, the more parameters are analyzed, enhancing the model's behavioral representation. However, to maintain computational efficiency, the default number of key parameters is set to $k=1,000$.


\subsection{Case Study}

\begin{figure}[!t]
\captionsetup[subfloat]{captionskip=5pt, font=scriptsize }
\centering
\subfloat[Simple Normal Sample]{\includegraphics[width=0.16\textwidth]{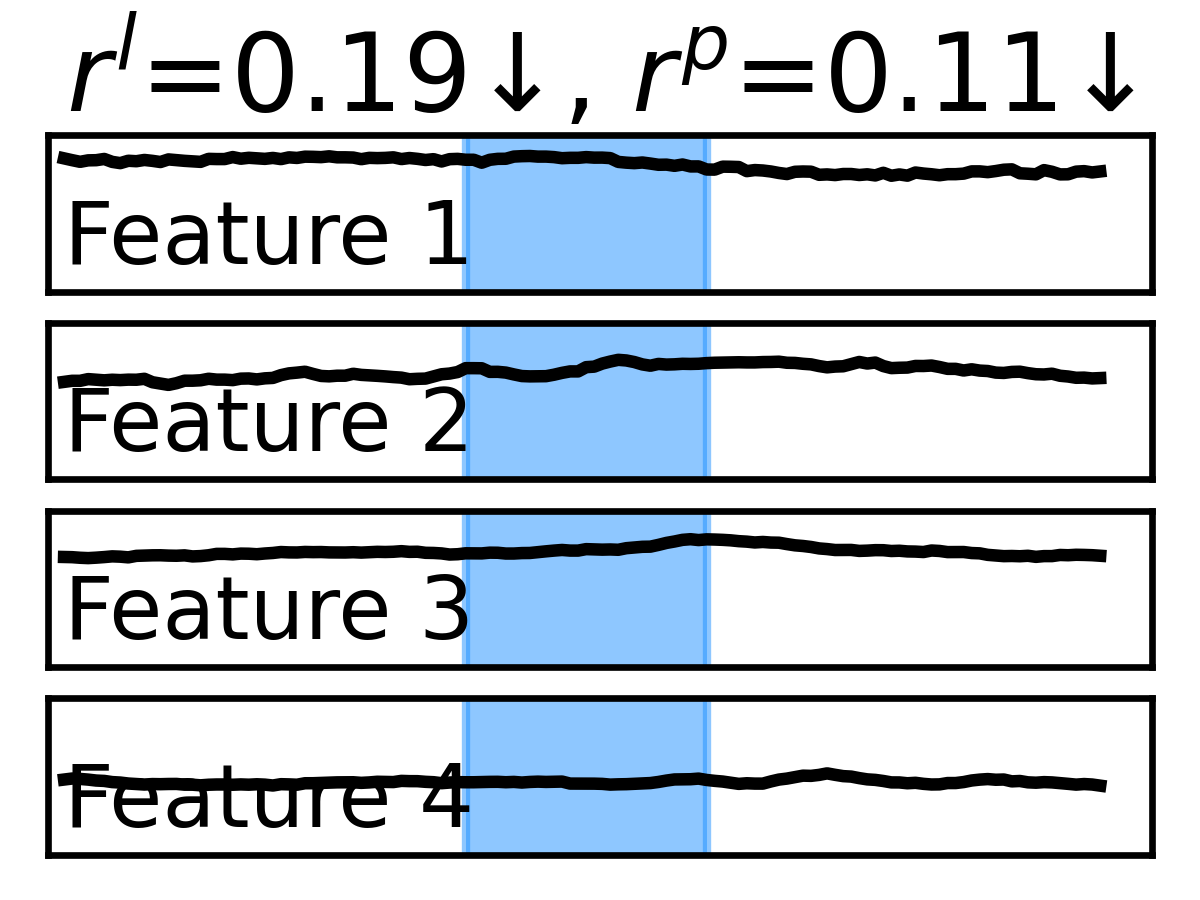}\label{fig:normal}}
\subfloat[Hard Sample]{\includegraphics[width=0.16\textwidth]{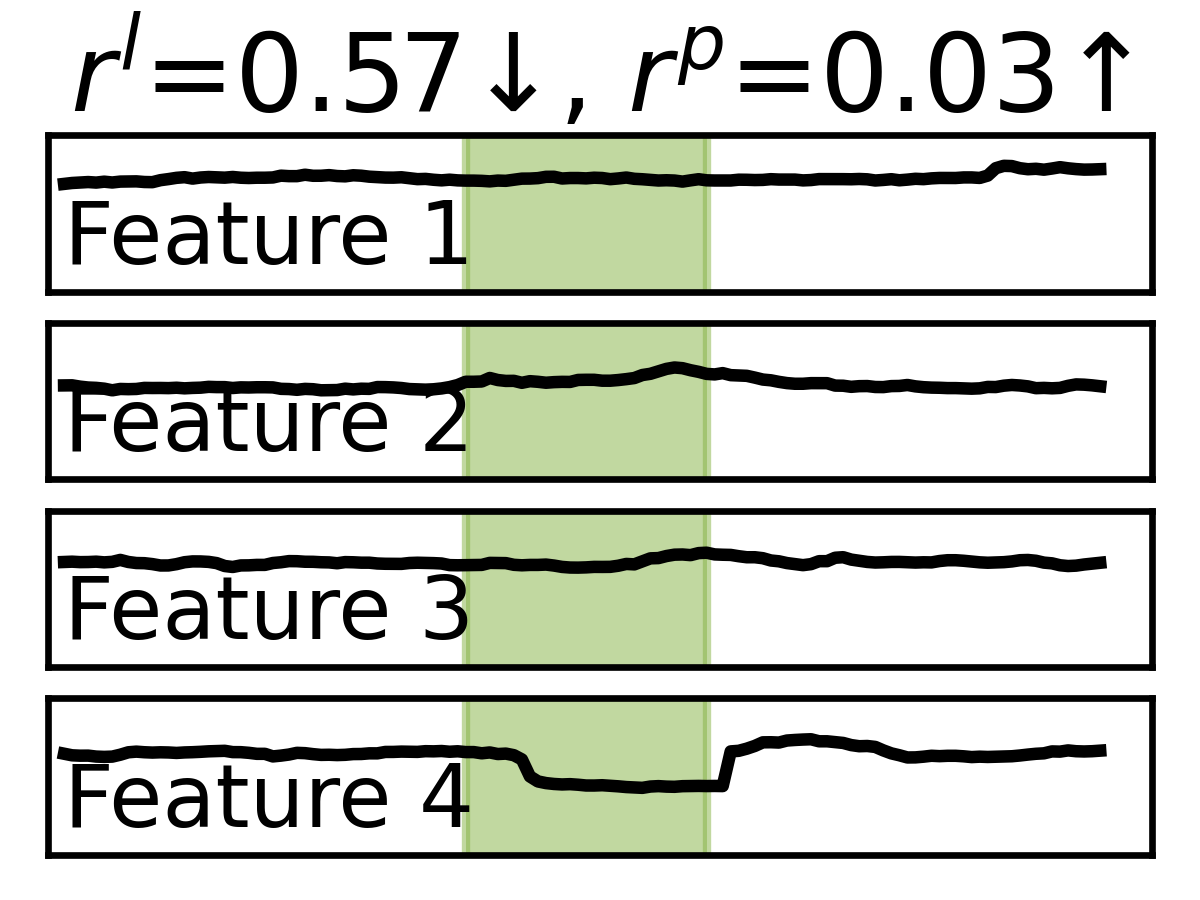}\label{fig:hard}}
\subfloat[Anomaly Contamination]{\includegraphics[width=0.16\textwidth]{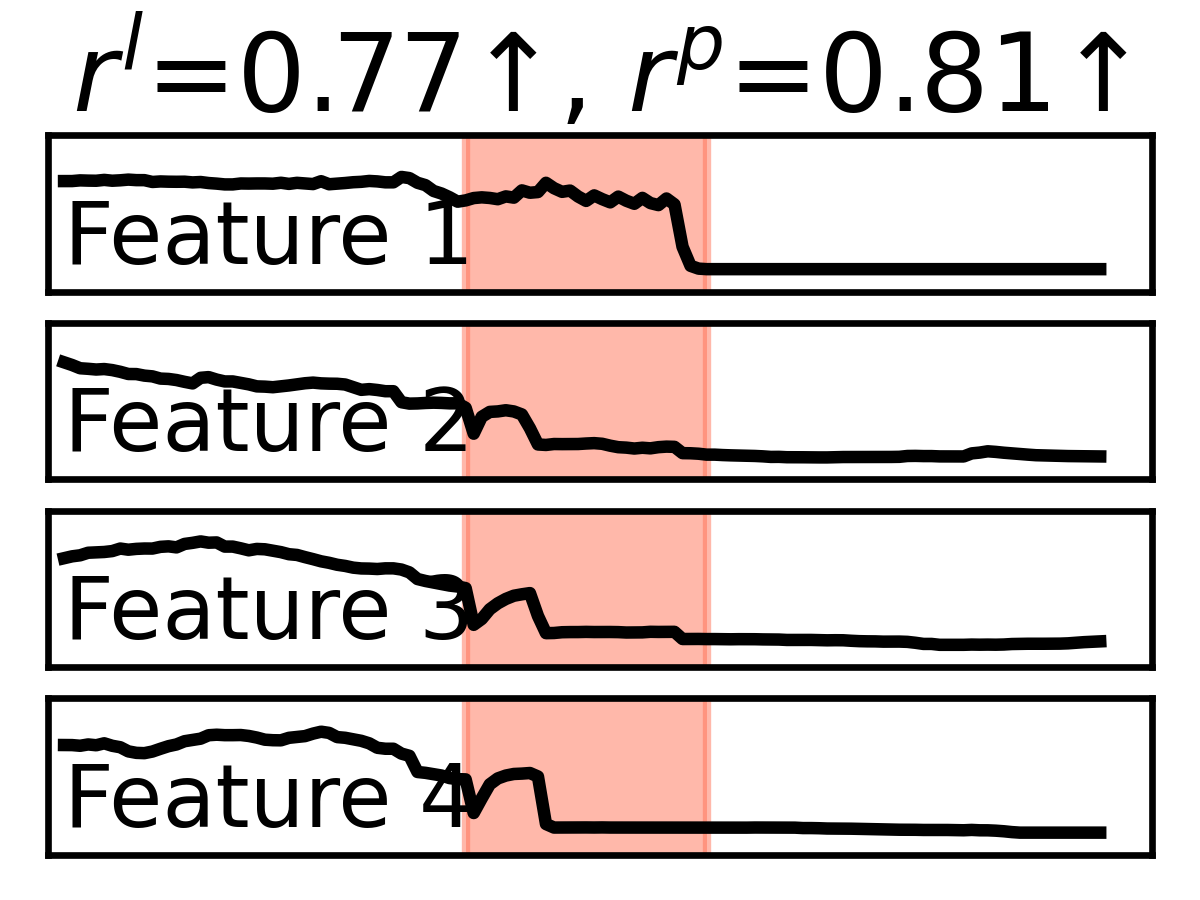}\label{fig:abnormal}}
\caption{Case study.
The parameter and loss behavior rewards for various sample types. $r^l$ indicates the loss behavior reward. $r^p$ indicates the parameter behavior reward.   
} 
\label{fig:case}
\end{figure}

This experiment delves into the efficacy of the behavior rewards we proposed, which is designed to distinguish between various categories of samples. \cref{fig:case} provides a visual representation of the parameter and loss behavior rewards allocated to simple normal samples, HS and AC, respectively. Both AC and HS are inherently challenging for learning algorithms, which is reflected in the high loss behavior rewards they receive. However, these two types of samples can be differentiated based on the specific parameter behavior rewards they are assigned. In contrast, simple normal samples consistently receive low rewards across both metrics. The experimental findings underscore that the two types of behavior rewards exhibit distinct preferences for different sample types, thereby enabling effective discrimination between them. 


\subsection{Ablation Study}



This study addresses two key issues: (\textbf{Q1})~Is the dual-dimensional reward of parameter and loss behaviors beneficial? (\textbf{Q2})~Is the proposed data augmentation method effective?  
We first design two simplified versions: \textbf{w/}~Param and \textbf{w/}~Loss, which distinguish AC and HS only based on the parameter behavior and the loss behavior, respectively. 
We design another three ablated variants: \textbf{w/}~Clus applies the clustering method to the feature space of parameter and loss behavior construction. It can validate the effectiveness of reinforcement learning.
\textbf{w/o}~Exp and \textbf{w/o}~Del, which respectively exclude the expansion and deletion operation. These two variants can validate the effectiveness of the adaptive sliding window module.

The results in \cref{tab:Ablation} show PLDA outperforming these variants, validating the roles of the key designs of PLDA in managing TSAD in contamination.
Regarding Q1, relying solely on either parameter or loss behavior typically results in diminished overall performance, with a 1.7\% reduction when depending solely on parameter behavior and a 3.4\% decrease when focusing only on loss behavior. Though slight performance perturbations of 0.2\%-0.6\% are observed, relying on a single indicator alone generally underperforms across most datasets. This underscores the significance of using both indicators, as a single indicator alone cannot adequately differentiate between varied data types. 


For Q2, clustering reduces performance by an average of 7.3\%. As TSAD is unsupervised with contaminated training data, classification is not applicable. Besides, clustering is ineffective due to unclear data boundaries and the need for manual category matching. In contrast, PLDA automatically determines data categories without sensitive hyperparameters, offering better performance. Excluding expansion results in a 6.6\% performance drop, showing the benefit of up-sampling HS. The exclusion of deletion has a 0.9\% impact, demonstrating its role in data augmentation. This confirms the adaptive sliding window module's effectiveness in handling diverse data types.

Based on the above comparison results, the contribution of the proposed parameter behavior and the data augmentation method are validated and quantitatively measured.

\begin{table}
	\setlength{\tabcolsep}{0.3pt}
	\renewcommand{\arraystretch}{1.3}
	\caption{
		F1-scores of PLDA and its ablation variants in \%. Positive gains in \textbf{bold}. Group 1 focuses on the dual-dimensional reward of parameter and loss behaviors; Group 2 focuses on the proposed data augmentation method, including the reinforcement learning framework and the adaptive sliding window module.
	}
	\label{tab:Ablation}
	\centering
	\scalebox{0.95}{
		\begin{tabular}{lcccccc}
			\toprule[1.5pt]
			&      & \multicolumn{2}{c}{\textbf{Group 1}} & \multicolumn{3}{c}{\textbf{Group 2}} \\
			\addlinespace[-\aboverulesep] 
			\cmidrule(r{0.2cm}){3-4} 
			\cmidrule{5-7} 
			\addlinespace[\belowrulesep] 
			
			\textbf{Data}                    & PLDA       & \textbf{w/} Param     & \textbf{w/} Loss    & \textbf{w/} Clus & \textbf{w/o} Exp      & \textbf{w/o} Del  \\
			\hline
			ASD                     & 85.2 & 82.6 (\textbf{3.2\%})  & 83.2 (\textbf{2.4\%}) & 82.1 (\textbf{3.8\%}) & 78.9 (\textbf{7.9\%})  & 85.1 (\textbf{0.1\%})  \\
			MSL                     & 89.0 & 89.4 (\textbf{0.0\%})  & 88.1 (\textbf{1.0\%})  & 88.5 (\textbf{0.5\%})& 86.8 (\textbf{2.5\%})  & 88.1 (\textbf{1.1\%})  \\
			SMAP                    & 71.1 & 70.3 (\textbf{1.1\%})  & 70.9 (\textbf{0.3\%})  & 70.7 (\textbf{0.5\%})& 70.9 (\textbf{0.2\%})  & 71.0 (\textbf{0.1\%})  \\
			SMD                     & 81.6 & 81.3 (\textbf{0.4\%})  & 81.3 (\textbf{0.4\%})  & 76.9 (\textbf{6.1\%})& 80.8 (\textbf{1.1\%})  & 81.8 (-0.2\%) \\
			SWaT                    & 84.1 & 84.0 (\textbf{0.1\%})  & 83.9 (\textbf{0.2\%})  & 82.1 (\textbf{2.5\%}) & 83.9 (\textbf{0.2\%})  & 84.0 (\textbf{0.1\%})  \\
			PUMP                    & 90.5 & 90.4 (\textbf{0.1\%})  & 90.3 (\textbf{0.2\%})  & 87.7 (\textbf{3.1\%}) & 89.9 (\textbf{0.7\%})  & 90.4 (\textbf{0.1\%})  \\
			DSADS                   & 97.8 & 96.3 (\textbf{1.6\%})  & 90.0 (\textbf{8.7\%})  & 96.9 (\textbf{0.9\%})& 89.3 (\textbf{9.6\%})  & 97.4 (\textbf{0.4\%}) \\
			Fault                   & 98.9 & 88.8 (\textbf{11.3\%}) & 85.0 (\textbf{16.3\%}) & 65.1 (\textbf{52.0\%})& 62.1 (\textbf{59.3\%}) & 91.4 (\textbf{8.1\%})  \\
			Gait                    & 95.4 & 95.9 (-0.5\%) & 90.8 (\textbf{5.1\%}) & 86.4 (\textbf{10.4\%})& 87.1 (\textbf{9.5\%})  & 96.0 (-0.6\%)  \\
			Heart             & 80.2 & 80.6 (\textbf{0.0\%})  & 80.4 (-0.2\%) & 76.22 (\textbf{5.2\%})& 80.4 (-0.2\%) & 79.6 (\textbf{0.8\%})  \\
			\hline
			Avg & 87.4  & 85.9 (\textbf{1.7\%})  & 84.4 (\textbf{3.4\%}) & 81.5 (\textbf{7.3\%})& 81.0 (\textbf{6.6\%})  & 86.5 (\textbf{0.9\%}) \\
			
			\toprule[1.5pt]
	\end{tabular}}
\end{table}

\section{Limitations and Future Directions}\label{app:future}
We present a dual parameter-loss data augmentation method to mitigate training set contamination issues in unsupervised TSAD.
The core advancement of our work is the formulation of a parameter behavior function.
This innovative function enables a nuanced description and explanation of neural network behaviors during anomaly detection.


However, a limitation exists in computing parameter behavior values: calculating the gradient for each parameter can be exceedingly computationally expensive. We design some details to improve the efficiency of the calculation, such as calculating only the critical $k$ parameters. \cref{sec:dataused} also proves that using the adaptive sliding window module in PLDA can significantly reduce the size of the training set, thereby alleviating this limitation. \cref{sec:scalability} verifies this assumption. Nonetheless, this issue may still be unavoidable when dealing with huge model network structures.

Looking ahead, we aim to delve into four promising directions:
\textbf{(1)}~Optimize the calculation method of parameter behavior and reduce the computational complexity.
\textbf{(2)}~Leveraging behavior values as direct anomaly indicators for better anomaly discrimination.
\textbf{(3)}~Expanding PLDA's utility to various data types, such as image and tabular data, by refining our approach.
\textbf{(4)}~Utilizing behavior values for model evaluation and selection, promoting models with consistent behavior as a new, label-independent evaluation standard.

\section{Conclusion}\label{sec:Conclusion}
This paper introduces PLDA, a dual parameter-loss data augmentation method, to overcome the challenges of unsupervised time series anomaly detection in contamination. 
Our work highlights the limitations of current methods in handling anomaly contaminations and hard samples.
Innovatively, we introduce a parameter behavior function and propose a dual-dimensional metric to identify sample types. 
PLDA utilizes reinforcement learning to iteratively identify and address hard samples and anomaly contaminations during the TSAD model's training phase. It serves as an additional step to augment the detection performance in a plug-and-play manner.
The efficacy of PLDA is validated through extensive experiments.
Looking ahead, PLDA paves new avenues for understanding neural network behaviors in anomaly detection, suggesting promising future explorations.

\section*{Acknowledgments}
The work was supported by the National Natural Science Foundation of China under Grant~62406328 and the Foundation of National University of Defense Technology under Grant~24-ZZCX-JDZ-07.

\appendix

\section*{Proof of \cref{the:dtheta}.}\label{app:theA}
Inspired by the work~\cite{koh2017understanding}, we provide a theoretical derivation of the parameter behavior function. 

\begin{proof}
    Given a training set $\mathcal{S} = \{\mathbf{s}_1, \mathbf{s}_2, \ldots, \mathbf{s}_n\}$ where  the weight of all samples are $1$, its empirical risk $E(\theta)$ is:
\begin{equation}
    E(\theta) = \frac{1}{n}\sum\limits_{i=1}^n L(\mathbf{s}_i, \theta).
\end{equation}

Disturbing a sample $\mathbf{s}$ with a small weight $\epsilon$, we get the new training set:
\begin{equation}
    \mathcal{S}'=\mathcal{S}\cup\epsilon\mathbf{s}=\{\mathbf{s}_1, \mathbf{s}_2, \ldots, \mathbf{s}_n, \epsilon\mathbf{s}\},
\end{equation}

in which the weight of sample $s$ becomes $1+\epsilon$. The optimized parameter $\hat{\theta}_{\epsilon,\mathbf{s}}$ becomes:
\begin{equation}
\hat{\theta}_{\epsilon,\mathbf{s}} = \mathop{\arg\min}\limits_{\theta}\left[\frac{1}{n} \sum\limits_{i=1}^{n}L(\mathbf{s}_i, \theta)+\epsilon L(\mathbf{s},\theta)\right].
\end{equation}

Then we define $\Delta\epsilon=\hat{\theta}_{\epsilon,s}-\hat{\theta}$ to evaluate the change of $\theta$. Note that $\hat{\theta}$ is the minimization result of empirical risk, and it is independent of $\epsilon$. Then we get:
\begin{equation}\label{eq:delta}
    \frac{\mathrm{d}\Delta\epsilon}{\mathrm{d}\epsilon}=\frac{\mathrm{d}(\hat{\theta}_{\epsilon, \mathbf{s}}-\hat{\theta})}{\mathrm{d}\epsilon}=\frac{\mathrm{d}\hat{\theta}_{\epsilon, \mathbf{s}}}{\mathrm{d}\epsilon}-\frac{\mathrm{d}\hat{\theta}}{\mathrm{d}\epsilon}=\frac{\mathrm{d}\hat{\theta}_{\epsilon,\mathbf{s}}}{\mathrm{d}\epsilon}.
\end{equation}

Because $\hat{\theta}_{\epsilon, \mathbf{s}}$ is the minimization result of \cref{eq:theta}, it satisfies the first derivative condition, that is, the first derivative with respect to $\theta$ is $0$:
\begin{equation}\label{eq:firstd}
    0 = \nabla_{\theta}E(\hat{\theta}_{\epsilon, \mathbf{s}})+\epsilon \nabla_{\theta}L(\mathbf{s}, \hat{\theta}_{\epsilon,\mathbf{s}}).
\end{equation}

When $\epsilon$ approaches $0$, $\hat{\theta}_{\epsilon, \mathbf{s}}$ also approaches $\hat{\theta}$. We perform a Taylor expansion of first order on the right side of \cref{eq:firstd}, that is, expand $\hat{\theta}_{\epsilon, \mathbf{s}}$ near $\hat{\theta}$. Then we obtain:
\begin{equation}
    \begin{aligned}
    0 \approx & \left[\nabla_{\theta} E(\hat{\theta}) + \epsilon\nabla_{\theta}L(\mathbf{s}, \hat{\theta})\right] \\
    & + \left[\nabla^2_{\theta}E(\hat{\theta}) + \epsilon\nabla^2_{\theta}L(\mathbf{s},\hat{\theta})\right] \Delta\epsilon.
    \end{aligned}
\end{equation}
where we omit the $R(\hat{\theta}_{\epsilon, \mathbf{s}})$ terms. 

Solving for $\Delta\epsilon$, then we get:
\begin{equation}\label{eq:deplison}
    \begin{aligned}
    \Delta\epsilon \approx & -\left[\nabla^2_{\theta}E(\hat{\theta}) + \epsilon\nabla^2_{\theta}L(\mathbf{s},\hat{\theta})\right]^{-1} \\
    & \times \left[\nabla_{\theta} E(\hat{\theta}) + \epsilon\nabla_{\theta}L(\mathbf{s}, \hat{\theta})\right].
    \end{aligned}
\end{equation}

$\hat{\theta}$ minimizes $E$, which means $\nabla_{\theta}E(\hat{\theta})=0$. Keeping only the $O(\epsilon)$ terms, we have:
\begin{equation}
    \Delta\epsilon \approx -\nabla^2_{\theta}E(\hat{\theta})^{-1}\nabla_{\theta}L(\mathbf{s}, \hat{\theta})\epsilon.
\end{equation}

Combining \cref{eq:delta}, relationship between parameter change is finally obtained:
\begin{equation}
    \left.\frac{\mathrm{d}\hat{\theta}_{\epsilon, \mathbf{s}}}{\mathrm{d}\epsilon}\right|_{\epsilon=0}=-H_{\hat{\theta}}^{-1}\nabla_{\theta}L(\mathbf{s}, \hat{\theta}),
\end{equation}
in which $H_{\hat{\theta}}=\nabla_{\theta}^2E(\hat{\theta})$ is the Hessian matrix. 
\end{proof}

\section*{Proof of \cref{the:lt}}\label{app:theB}
Inspired by the work~\cite{xu2018understanding}, we provide the proof of parameter behavior function effectiveness.

\begin{proof}
Given a neural network with one hidden layer with $N$ nodes and a tanh function $\sigma(\mathbf{s})$ as the activation function. It can be expressed as:
\begin{equation}
    D(\mathbf{s})=\sum\limits_{j=1}^N \alpha_j\sigma(\omega_j\mathbf{s}+\beta_j),
\end{equation}
in which $\alpha_j, \omega_j,\beta_j\in\mathbbm{R}$ are parameters of the DNN. 

The Fourier transform of $D(\mathbf{s})$ is:
\begin{align}\label{eq:D}
    \mathcal{F}(D(\mathbf{s}))(f) = &\sum\limits_{j=1}^N\alpha_j \left[
    \sqrt{\frac{\pi}{2}}\delta(f) + \right. \\
    &\left. \sqrt{\frac{\pi}{2}}\frac{\mathrm{i}}{|\omega_j|}
    \frac{\mathrm{e}^{-\frac{\mathrm{i}\beta_j f}{\omega}}}
    {\mathrm{e}^{\frac{\pi f}{2\omega_j}}-\mathrm{e}^{-\frac{\pi f}{2\omega_j}}}
    \right].
\end{align}
Assume $\frac{\pi f}{\omega_j}\gg 0$ and $\omega_j>0$, then \cref{eq:D} is reduced to:
\begin{equation}
    \mathcal{F}(D(\mathbf{s}))(f)=\sum\limits_{j=1}^N\alpha_j \left[\sqrt{\frac{\pi}{2}}\frac{\mathrm{i}}{\omega_j}\mathrm{e}^{-\frac{\pi f}{2\omega_j}-\frac{\mathrm{i}\beta_j f}{\omega_j}} \right].
\end{equation}
Write the difference between DNN output and the original sample at each frequency as:
\begin{equation}
    \begin{aligned}
    J(\mathbf{s}(f), \theta)
    &=\mathcal{F}(D(\mathbf{s}))(f)-\mathcal{F}(\mathbf{s})(f)
    &=A(f)\mathrm{e}^{\mathrm{i}\phi(f)},
    \end{aligned}
\end{equation}
in which $\theta_j\in\{\alpha_j, \beta_j, \omega_j\}$. $A(f),\phi(f)\in[-\pi,\pi]$ are the amplitude and phase of $J(\mathbf{s}(f), \theta)$, respectively. The loss at frequency $f$ is $L(\mathbf{s}(f), \theta)=\Vert J(\mathbf{s}(f), \theta)\Vert_2$. The total loss is expressed as:
\begin{equation}
    L(\mathbf{s}, \theta)=\sum\limits_f L(\mathbf{s}(f), \theta).
\end{equation}
Then we can calculate the gradient of each parameters: 
\begin{enumerate}
    \item For the gradient of $\alpha$:
\begin{align}
    \frac{\partial L(\mathbf{s}(f), \theta)}{\partial \alpha_j} = & \, \overline{J(\mathbf{s}(f), \theta)}\frac{\partial J(\mathbf{s}(f), \theta)}{\partial \alpha_j} \nonumber \\
    & + J(\mathbf{s}(f), \theta)\frac{\partial \overline{J(\mathbf{s}(f), \theta)}}{\partial \alpha_j}.
\end{align}

Since
\begin{equation}
    \begin{aligned}
    \frac{\partial J(\mathbf{s}(f), \theta)}{\partial\alpha_j} 
    &= \frac{\partial \mathcal{F}(D(\mathbf{s})(f))}{\partial\alpha_j} \\
    &= \sqrt{\frac{\pi}{2}}\frac{\mathrm{i}}{\omega_j}\mathrm{e}^{-\frac{\pi f}{2\omega_j}-\frac{\mathrm{i}\beta_j f}{\omega_j}},
    \end{aligned}
\end{equation}
then
\begin{equation}
    \begin{aligned}
    \frac{\partial L(\mathbf{s}(f), \theta)}{\partial\alpha_j} 
    &= A(f)\sqrt{\frac{\pi}{2}}\frac{\mathrm{i}}{\omega_j} 
    \left[
    \mathrm{e}^{-\frac{\pi f}{2\omega_j}-\frac{\mathrm{i}\beta_j f}{\omega_j}-\mathrm{i}\phi(f)}
    \right. \\
    & \quad \left.
    - \mathrm{e}^{-\frac{\pi f}{2\omega_j}+\frac{\mathrm{i}\beta_j f}{\omega_j}+\mathrm{i}\phi(f)}
    \right] \\
    &= A(f)\sqrt{\frac{\pi}{2}}\frac{\mathrm{i}}{\omega_j} 
    \mathrm{e}^{-\frac{\pi f}{2\omega_j}+\frac{\mathrm{i}\beta_j f}{\omega_j}+\mathrm{i}\phi(f)} \\
    & \quad \times
    \left[
    \mathrm{e}^{-\frac{2\mathrm{i}\beta_j f}{\omega_j}-2\mathrm{i}\phi(f)} - 1
    \right].
    \end{aligned}
\end{equation}

Denote
\begin{subequations}
    \begin{align}
        I_0&=\sqrt{\frac{\pi}{2}}\mathrm{e}^{\frac{\mathrm{i}\beta_j f}{\omega_j}+\mathrm{i}\phi(f)},\\
        I_1&=\mathrm{e}^{\frac{-2\mathrm{i}\beta_j f}{\omega_j}-2\mathrm{i}\phi(f)}.
    \end{align}
\end{subequations}
Then we have
\begin{equation}
\label{eq:alpha}
    \frac{\partial L(\mathbf{s}(f), \theta)}{\partial\alpha_j}=I_0(I_1-1)\frac{\mathrm{i}}{\omega_j}A(f)\mathrm{e}^{-\frac{\pi f}{w\omega_j}}.
\end{equation}

\item For the gradient of $\beta$,
\begin{equation}
    \begin{aligned}
        \frac{\partial J(\mathbf{s}(f), \theta)}{\partial \beta_j}&=\frac{\partial \mathcal{F}(D(\mathbf{s}))(f)}{\partial \beta_j}\\
        &=\sqrt{\frac{\pi}{2}}\frac{\alpha_j}{\omega_j^2}\mathrm{e}^{-\frac{f}{\omega_j}(\mathrm{i}\beta+\frac{\pi}{2})}f.
    \end{aligned}
\end{equation}
Then we get 
\begin{equation}
\label{eq:beta}
    \frac{\partial L(\mathbf{s}(f), \theta)}{\partial \beta_j}=(I_1-1)I_0\frac{\alpha_j f}{\omega_j^2}A(f)\mathrm{e}^{-\frac{\pi f}{2\omega_j}}.
\end{equation}

\item For the gradient of $\omega$:
\begin{equation}
    \begin{aligned}
    \frac{\partial J(\mathbf{s}(f), \theta)}{\partial\omega_j} &= \frac{\partial \mathcal{F}(D(\mathbf{s}))(f)}{\partial\omega_j} \\
    &= \sqrt{\frac{\pi}{2}} \frac{\alpha_j}{2\omega_j^3} \mathrm{e}^{-\frac{ f}{\omega_j}(\beta_j+\frac{\pi}{2})} \\
    &\quad \times (\mathrm{i}\pi f - 2\mathrm{i}\omega_j - 2\beta_j f),
    \end{aligned}
\end{equation}
then
\begin{align}
    \frac{\partial L(\mathbf{s}(f), \theta)}{\partial\omega_j} = & \, A(f)I_0\frac{\alpha_j}{2\omega^3}\mathrm{e}^{-\frac{\pi f}{2\omega_j}} \nonumber \\
    & \times \left[ I_1(\mathrm{i}\pi f - 2\mathrm{i}\omega_j - 2\beta_j f) \right. \nonumber \quad \\
    &\left. + (-\mathrm{i}\pi f + 2\mathrm{i}\omega_j - 2\beta_j f) \right].
\end{align}
Denote
\begin{equation}
    I_2=I_1(\mathrm{i}\pi f-2\mathrm{i}\omega_j-2\beta_j f)+(-\mathrm{i}\pi f+2\mathrm{i}\omega_j-2\beta_j f).
\end{equation}
Then we get
\begin{equation}
\label{eq:omega}
    \frac{\partial L(\mathbf{s}(f), \theta)}{\partial\omega_j}=I_0 I_2\frac{\alpha_j}{2\omega_j^3}A(f)\mathrm{e}^{-\frac{\pi f}{2\omega_j}}.
\end{equation}

\end{enumerate}

To sum up, the absolute contribution from frequency $f$ to the total amount at $\theta$ is given by
\begin{equation}
    \left|\frac{\partial L(\mathbf{s}(f), \theta)}{\theta_j}\right|=A(f)\mathrm{e}^{-\left|\frac{\pi f}{2\omega_j}\right|}K(\theta_j, f).
\end{equation}
$K(\theta_j, f)$ is a function of $\theta_j$ and $f$, which can be find in \cref{eq:alpha,eq:beta,eq:omega}.
Since a small $\omega_j$ and $\mathrm{e}^{-\left|\frac{\pi f}{2\omega_j}\right|}$ would dominate $K(\theta_j, f)$, the function can be written as:
\begin{equation}
    P(\mathbf{s}(f), \theta_j)=\left|\frac{\partial L(\mathbf{s}(f),\theta)}{\partial\theta_j}\right|\approx A(f)\mathrm{e}^{-\left|\frac{\pi f}{2\omega_j}\right|}.
\end{equation}
\end{proof}

\section*{Proof of Expansion Action Validity}\label{app:exp}

Time series data frequently displays periodicity, and using a whole period for model training improves pattern recognition. Yet, sliding windows can split a period into two windows. The period's start might be at any position, necessitating any sample expandable with limited extensions. We demonstrate that the extended action of PLDA can meet the above requirement.

\begin{proof}
Set the initial sample $\mathbf{s}_0=\mathbf{s}^i$, where $\mathbf{s}^i=\langle \mathbf{x}^i, \mathbf{x}^{i+1}, \ldots, \mathbf{x}^{i+w-1}\rangle$ indicates a sample starting from $\mathbf{x}^i$ with length $w$. 
After $k$ actions, we obtain the expanded sample:
\begin{equation}
    \mathbf{s}_k = \mathbf{s}^{i-k_1 w_1+k_2 w_2},
\end{equation}
where $k=k_1+k_2, k\in\mathbbm{N}^*,k_1,k_2\in\mathbbm{N}$. This expansion action consists of $k_1$ steps of forward expansion with a length of $w_1$ and $k_2$ steps of backward expansion with a length of $w_2$.

We aim to prove that PLDA can expand $\mathbf{s}^{i+p}$ into the training set for any $p \in \mathbbm{N}^+$ in $k$ actions. It is evident that if $\mathbf{s}^{i+1}$ is expandable, repeating the process $p$ times ensures $\mathbf{s}^{i+p}$ is also expandable. Assume, for the sake of argument, that $p=1$. Then we get:
\begin{equation}
\label{eq:condition}
       -k_1 w_1+k_2 w_2=1, \quad
       \text{s.t.}\quad \begin{cases}
           w_1+w_2=w,\\
           k_1+k_2=k,\\
           w_1, w_2, k\in \mathbbm{N}^*,\\
           k_1, k_2 \in \mathbbm{N}.
       \end{cases}
\end{equation}
We use proof by contradiction to show that $w_1$ and $w_2$ must be coprime.
Assuming \cref{eq:condition} is satisfied, when $w_1$ and $w_2$ are not coprime. Set the greatest common factor of them as $c\neq1$ (i.e., $w_1 = ac$ and $w_2 = bc$, $a,b,c\in\mathbbm{N}^*$). Then:
\begin{equation}
       -k_1 w_1+k_2 w_2=(-k_1 a+k_2 b)c, \quad
       \text{s.t.}\quad \begin{cases}
           c \neq 1,\\
           a,b,c\in\mathbbm{N}^*,\\
           k_1,k_2\in\mathbbm{N}.
       \end{cases}
\end{equation}
Since $-k_1 a+k_2 b \geq 1$ and $c>1$, their product cannot equal $1$, which is contradictory with the \cref{eq:condition}. Therefore, the requirement is met only when $w_1$ and $w_2$ are mutually prime.

In PLDA, when $w$ is an odd number, we set $w_1=\frac{w-1}{2}$ and $w_2=\frac{w+1}{2}$, which are mutually prime. When $w$ is an even number, a pair of prime $w_1$ and $w_2$ also exists, according to Goldbach's Conjecture~\cite{wang2002goldbach}. In summary, PLDA can expand any sample into the training set.

\end{proof}





\end{document}